\documentclass{article}
\PassOptionsToPackage{table}{xcolor}
\usepackage[T1]{fontenc}
\usepackage{iclr2027_conference,times}

\usepackage{amsmath,amsfonts,bm}

\def\eqref#1{equation~\ref{#1}}

\def\1{\bm{1}}

\DeclareMathAlphabet{\mathsfit}{\encodingdefault}{\sfdefault}{m}{sl}
\SetMathAlphabet{\mathsfit}{bold}{\encodingdefault}{\sfdefault}{bx}{n}

\usepackage{booktabs,multirow,array}
\usepackage{hyperref}
\usepackage{url}
\usepackage{enumitem}
\usepackage{graphicx}
\usepackage[table]{xcolor}
\usepackage{caption}
\usepackage{fontawesome5}
\definecolor{githubblue}{HTML}{4183C4}

\title{%
\raggedright
\hyphenpenalty=10000
Zooming without Zooming v2: From Seeing Details to Connecting Evidence for Generalizable Multimodal Perception
}

\author{%
\parbox[t]{\dimexpr\textwidth-2\tabcolsep\relax}{%
\raggedright
\normalsize\bfseries
Lai Wei$^{1,3,*}$, Yatai Li$^{1,2,*}$, Liangbo He$^{2,*}$,
Jun Lan$^{2,\ddagger}$, Renjie Gu$^{5}$, Linhan Cao$^{1,2}$, \\
Junkai Lin$^{1}$, Huijia Zhu$^{2}$,
Weiqiang Wang$^{2}$, Linghe Kong$^{1}$, Yue Wang$^{3}$, \\
Zhuosheng Zhang$^{1}$, Weiran Huang$^{1,4,\dagger}$ \\[0.15em]
\normalfont\normalsize
$^{1}$School of Computer Science, Shanghai Jiao Tong University \quad $^{2}$Ant Group \\
$^{3}$Zhongguancun Academy \quad
$^{4}$Shanghai Innovation Institute \quad $^{5}$Fudan University
}}

\iclrfinalcopy

\newcommand{\ours}{\textsc{EvidenceConnect}}

\begin{document}
\maketitle
\lhead{Preprint}

\begingroup
\renewcommand{\thefootnote}{}
\footnotetext{%
\raggedright
\noindent
$^{*}$Equal contribution.
\quad $^{\ddagger}$Project Lead.\\
$^{\dagger}$Correspondence to Weiran Huang
(\texttt{weiran.huang@sjtu.edu.cn}).
}
\endgroup

\begin{figure*}[ht]
\centering
\includegraphics[
    width=0.95\textwidth,
    trim={0mm 1mm 6mm 8mm},
    clip
]{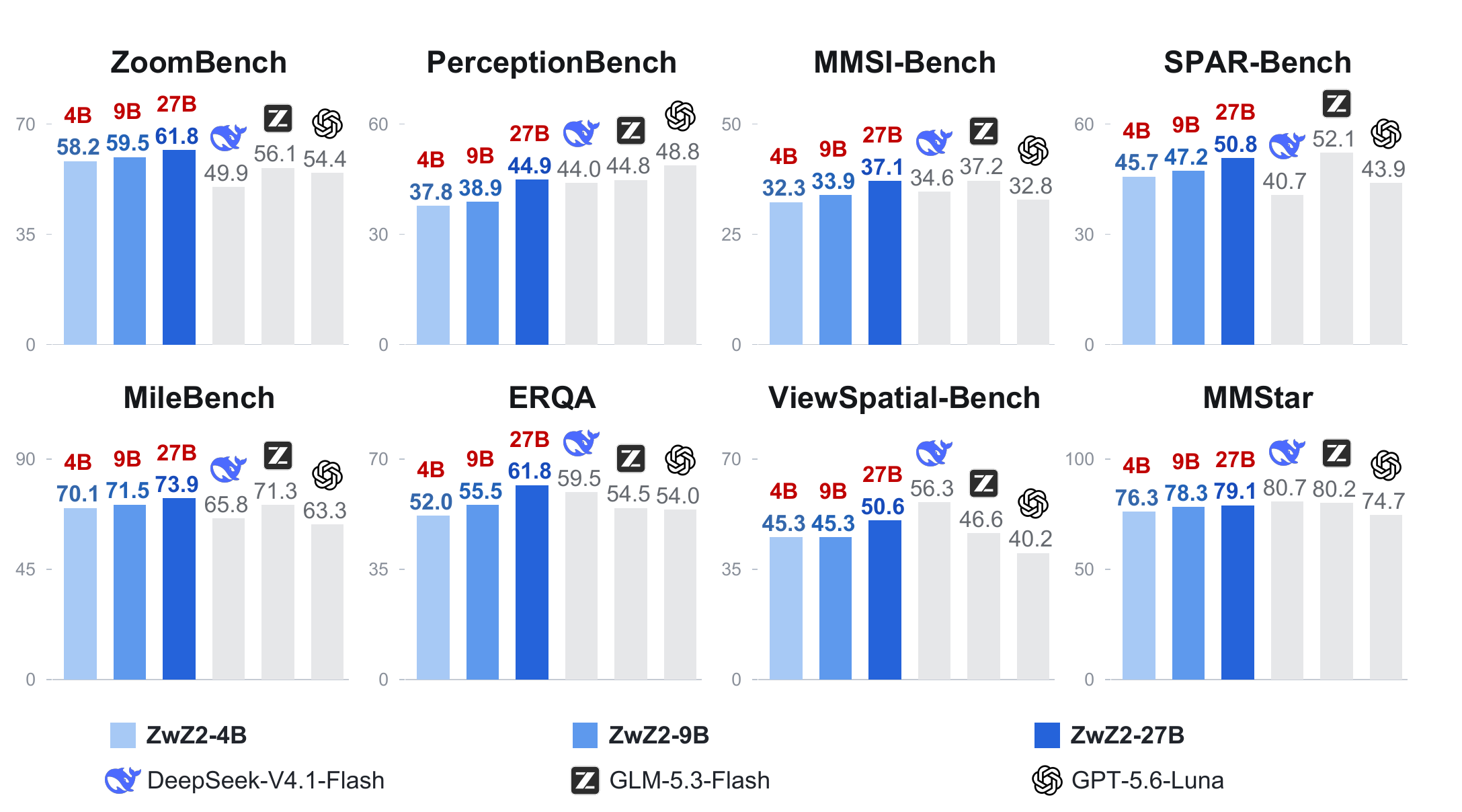}
\caption{Performance comparison of ZwZ2 and other advanced models.}
\label{fig:teaser}
\end{figure*}

\begin{center}

\footnotesize
  \hypersetup{hidelinks}
  \begin{tabular}{@{}c@{\hspace{0.35em}}l@{\hspace{0.6em}}l@{}}
    \makebox[1.3em][c]{\faGithub} &
    \textbf{Code:} &
    \href{https://github.com/inclusionAI/Zooming-without-Zooming-v2}{%
      \textcolor{githubblue}{%
        \texttt{inclusionAI/Zooming-without-Zooming-v2}%
      }%
    }\\[4pt]

    \makebox[1.3em][c]{%
      \raisebox{-0.15em}{%
        \includegraphics[height=1.15em]{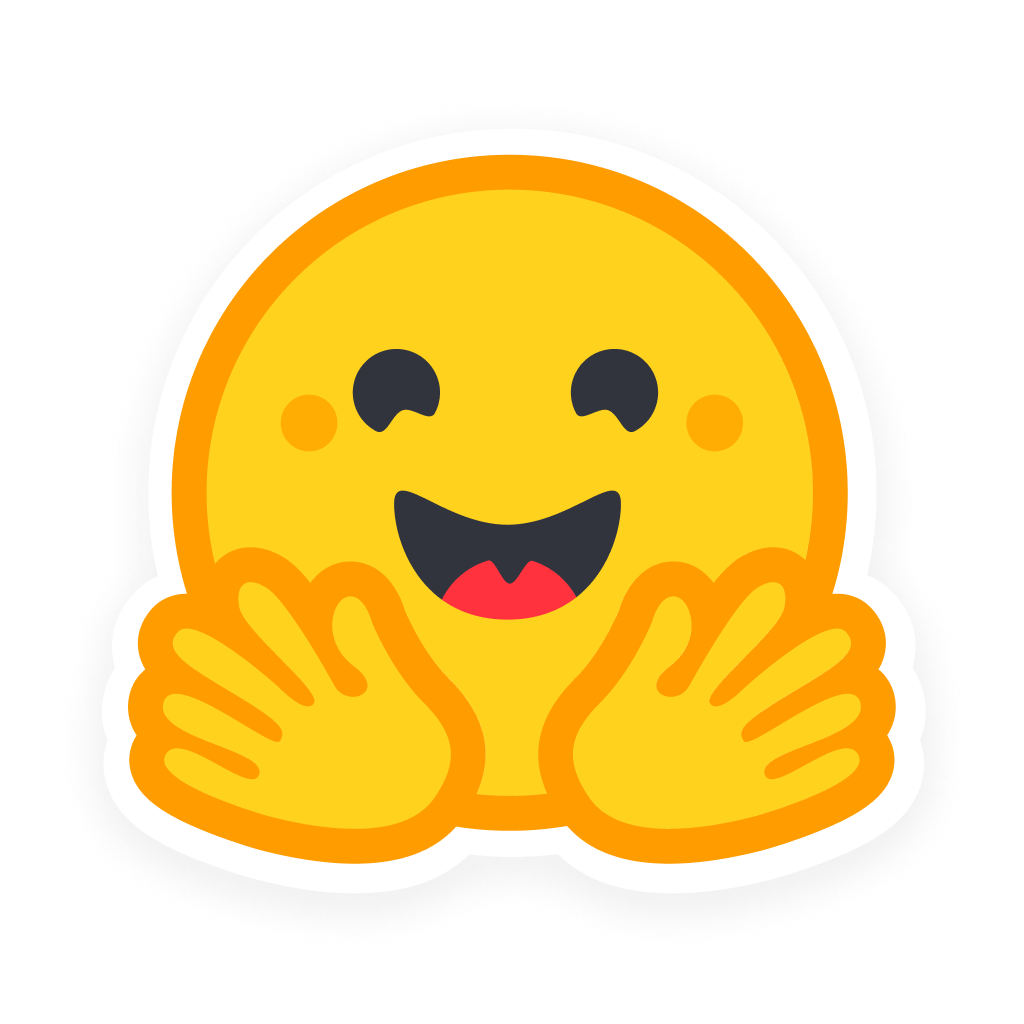}%
      }%
    } &
    \textbf{Hugging Face:} &
    \href{https://huggingface.co/collections/inclusionAI/zooming-without-zooming-v2}{%
      \textcolor{githubblue}{%
        \texttt{collections/inclusionAI/zooming-without-zooming-v2}%
      }%
    }
  \end{tabular}
\end{center}

\begin{abstract}
Multimodal perception forms the foundation of multimodal intelligence.
In particular, spatial perception and multimodal long-context understanding are two pivotal tasks that underpin embodied AI and multimodal agents. However, synthesizing targeted, high-quality training data for these tasks remains difficult and costly.
Our key insight is that these tasks share an underlying capability, which we call connecting evidence: identifying relevant visual facts, binding them to the correct entities, recovering their relations, and composing them into an answer.
Crucially, targeted training data are far easier to synthesize for this capability than for these tasks themselves.
To strengthen this capability, we propose \ours{}, an easy-to-implement data synthesis framework that uses zoomed views of related regions within a single image to generate questions requiring comparison, aggregation, or chaining of visual evidence.
To ensure quality, a reference answer is retained only when independent VLMs agree on it, and region-ablation tests discard questions whose answer survives the removal of selected regions.
We then train base VLMs on the resulting data with RLVR, obtaining Zooming without Zooming v2 (ZwZ2).
Although the synthesized data are not tailored to any downstream benchmark, ZwZ2 improves fine-grained and spatial perception, with further gains on embodied spatial reasoning, multimodal long-context understanding, multimodal agent tasks, and general visual reasoning (shown in Figure~\ref{fig:teaser}).
These results establish \ours{} as an effective framework for supervising evidence connection and suggest that strengthening this shared capability leads to more generalizable multimodal perception.

\end{abstract}

\section{Introduction}
\label{sec:intro}

Multimodal perception is the foundation of multimodal intelligence: before a multimodal model can reason or act, it must determine what is present in a scene. Recent progress on this front, such as multimodal Chain-of-Thought~\citep{zhang2023multimodal, xu2024llava, li-ma-2026-aim} and Thinking-with-Images~\citep{zheng2025deepeyes,zhang2025thyme}, has improved the ability of open-source vision-language models (VLMs) to perceive fine-grained visual content and use it in reasoning~\citep{wei2026zooming,wei2025advancing,li2026fewer}.

However, these advances do not reliably extend to tasks in which critical visual cues are dispersed across objects, relationships, frames, or dialogue turns. Two such tasks, spatial perception and multimodal long-context understanding, are particularly important because they underpin embodied AI and multimodal agents, respectively.
Spatial perception requires a model to recover how entities are arranged relative to one another and how several spatial relations compose~\citep{erqa2025,li2025viewspatial}. Embodied AI relies on it to ground navigation, manipulation, and viewpoint-dependent decisions in object locations and spatial relations~\citep{brohan2023rt,kim2024openvla,yang2026mmsi}. Multimodal long-context understanding requires a model to locate evidence across images and dialogue history, preserve entity identity, and bring that evidence to bear on the current question~\citep{wu2025visualhaystacks,li2024mmdu,wang2025mmlongbench,song2024milebench}. Multimodal agents rely on it to retrieve relevant past observations, resolve references across interactions, and integrate earlier evidence into current decisions~\citep{he2024webvoyager,su2026agentvista,li2025screenspot}.

Strengthening these capabilities, however, is constrained by the difficulty of building high-quality training data. 
For spatial perception, existing efforts mainly scale up region-grounded relational data~\citep{chen2024spatialvlm, zhang2026tvg}, but reliable labels require precise localization and relational grounding across multiple objects. Such labels are expensive to annotate by hand, while automated generators often hallucinate relations. 
For long-context understanding, existing multi-image and multi-turn instruction data~\citep{jiang2024mantis, li2024mmdu} concentrate on a few formats and task types, because diverse questions that remain answerable over long sequences are hard to construct automatically. 
Data for embodied spatial reasoning and multimodal agents are even harder to synthesize, since they additionally involve physical or interactive environments and multi-step trajectories. Consequently, neither these capabilities nor the applications built on them have a scalable and high-quality source of clean, verifiable supervision.

Our key insight is that these tasks build on a shared atomic capability of VLMs, which we call \emph{connecting evidence}: identifying task-relevant visual facts, binding each fact to the correct entity, recovering the relations among them, and integrating the evidence into an answer.
These tasks instantiate this capability differently: spatial perception connects objects within a scene through spatial relations, whereas long-context understanding, on which multimodal agents rely, connects facts across images and dialogue turns through entity identity and reference. 
Despite these differences in where the evidence resides and how it is linked, both tasks rely on the same underlying steps, so strengthening these steps can benefit both. Moreover, since the first step, identifying visual facts, is itself fine-grained perception, fine-grained perception benefits as well.

Crucially, training data that target this capability are far easier to construct and verify than data for the tasks themselves. To turn this insight into trainable supervision, we propose \ours{}, an easy-to-implement, fully automated framework that synthesizes visual questions requiring the model to connect evidence across regions.
Specifically, we select a group of related regions within a single image and synthesize a compositional question over them from \emph{privileged} zoomed views, which make small details visible but are used only during data construction.
Depending on the scene, we synthesize questions to compare, aggregate, or chain evidence across the regions.
Answering such questions thus exercises the same steps as the target tasks: relating regions spatially, including from a given viewpoint, parallels spatial perception and embodied spatial reasoning, whereas locating scattered regions, binding their facts to the right references, and integrating them parallels long-context understanding and multimodal agents.
To ensure data quality, we keep a question only if VLMs from different families agree on its answer, and we use region-ablation tests to discard questions whose answer survives the removal of selected regions.
Finally, each verified question is converted into multiple-choice form for rule-based rewards and paired with the full image, where colored outlines typically mark the selected regions to avoid ambiguous references, so the model must still perceive and connect the evidence by itself.

To evaluate whether this supervision generalizes beyond the synthesized questions, we train Qwen3.5-4B, Qwen3.5-9B, and Qwen3.6-27B~\citep{qwen3.5,qwen3.6-27b} with reinforcement learning with verifiable rewards (RLVR)~\citep{lambert2024tulu3,guo2025deepseekr1} using only our synthesized data, without supervised fine-tuning or benchmark-specific question formats. 
The resulting models, ZwZ2-4B/9B/27B, achieve substantial improvements in fine-grained and spatial perception as well as multimodal long-context understanding, with gains that further extend to embodied spatial reasoning, multimodal agent tasks, and general visual reasoning.
For example, ZwZ2-4B improves over its base model from 37.43 to 45.26 (+7.83) on ViewSpatial-Bench (spatial reasoning) and from 60.08 to 70.08 (+10.00) on MileBench (long-context understanding).
More broadly, its average over all 25 benchmarks increases from 49.41 to 54.79, and the corresponding averages improve from 51.63 to 56.59 for ZwZ2-9B and from 55.52 to 60.51 for ZwZ2-27B.
Notably, although ZwZ2 is trained only on single images, it also improves on multi-image long-context benchmarks and multimodal agent tasks. These results demonstrate that connecting evidence is indeed a generalizable training target for multimodal perception.

To summarize, our key contributions are the following:
\begin{itemize}[itemsep=0pt,topsep=0pt,leftmargin=0.5cm]
    \item We identify connecting evidence as an important atomic capability of VLMs, underlying fine-grained perception, spatial perception, and multimodal long-context understanding.
    \item We develop \ours{}, a structured and audited synthesis pipeline for single-image, multi-region compositional questions, with privileged region views and independent checks for answer correctness and evidence dependence.
    \item Experiments demonstrate that RLVR on the resulting synthetic data improves a broad range of perception benchmarks and further generalizes to many real-world tasks, including embodied spatial reasoning, multimodal agent tasks, and general visual reasoning.
\end{itemize}

\section{Related Work}

\textbf{Spatial perception.}~~
Spatial perception underpins embodied tasks such as robotic manipulation and embodied question answering~\citep{brohan2023rt,kim2024openvla,erqa2025}.
Existing approaches use large-scale spatial supervision~\citep{chen2024spatialvlm}, intermediate visual grounding~\citep{zhang2026tvg}, and progressive training~\citep{li2026spatialladder}.
Benchmarks and associated training resources examine perspective-taking, multi-image inference, and multi-hop composition~\citep{li2025viewspatial,yang2026mmsi}.
MultihopSpatial additionally studies grounded spatial training and its transfer to embodied tasks~\citep{lee2026multihopspatial}.
Euclid's Gift~\citep{lian2025euclid} demonstrates transfer from geometric surrogate tasks.
Similarly, we train on a surrogate task and evaluate its transfer, but our task is built from natural images and targets the evidence connection shared by spatial perception and long-context understanding, rather than spatial skills alone.

\textbf{Long-context multimodal understanding.}~~
Multimodal long-context understanding requires locating, tracking, and integrating evidence across long sequences of images and dialogue turns.
Existing efforts support these abilities through multi-image and multi-turn instruction tuning~\citep{jiang2024mantis,hu2024sparkles,li2024mmdu} and dialogue-fragment retrieval~\citep{bi2026f2rvlm}, while benchmarks assess long-context integration and memory~\citep{song2024milebench,wang2025mmlongbench,ren2026memlens}.
Controlled studies nevertheless identify persistent difficulties in aggregating distributed visual information~\citep{das2026more}.
We target evidence connection as a shared component of these tasks, constructing single-image questions that require identifying, binding, and composing facts across regions.
This setting makes the selected evidence easier to localize and audit, allowing us to investigate whether such supervision transfers to multimodal long-context understanding without training on long image sequences or dialogue histories.

\textbf{Multimodal synthetic data.}~~
Early resources provide annotated scene graphs and programmatic questions~\citep{krishna2017visualgenome,hudson2019gqa}. Later methods automate instruction generation~\citep{zhao2024genixer,zhang2025oasis}, evolution~\citep{luo2025mmevol}, and synthesis for unsupervised post-training~\citep{wei2026first}.
Zooming without Zooming~\citep{wei2026zooming} transfers single-region supervision to full images with cross-model answer checks.
CRIT~\citep{sung2026crit} and FM2DS~\citep{abaskohi2024fm2ds} synthesize cross-modal questions and filter unimodal shortcuts, while MM-CondChain~\citep{shen2026mm} verifies conditional programs and ReasonVQA~\citep{tran2025reasonvqa} composes external knowledge paths.
HopChain~\citep{wang2026hopchain} targets long-CoT robustness through instance-dependent chains with human-verified answers.
Unlike these pipelines, \ours{} builds each question on a coherent group of regions rather than a single region or a sequential chain, so its questions cover aggregation and closed-set comparison in addition to dependency chains.
Moreover, it automatically checks both the answer and its dependence on the selected regions, through cross-model agreement and region ablation, without human annotation.

\section{Boosting Multimodal Perception by Learning to Connect Visual Evidence}
\label{sec:method}

\begin{figure*}[t]
    \centering
    \includegraphics[
        width=0.99\textwidth,
        trim={8mm 0mm 8mm 0mm},
        clip
    ]{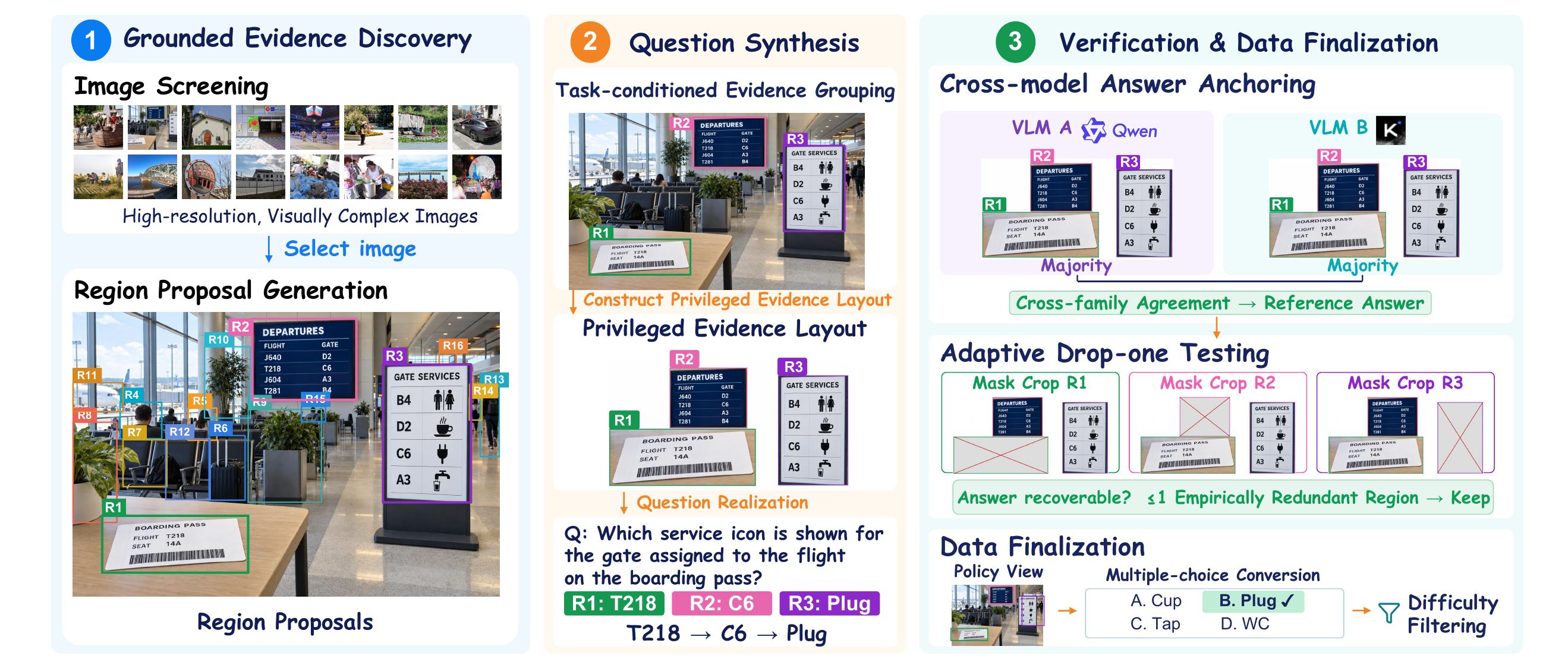}
    \caption{Overview of our multi-region question synthesis pipeline. We select related regions from a single image and construct a privileged crop view to generate compositional questions. The illustrative example connects a flight number on a boarding pass to its departure gate and then to a service icon (T218 $\rightarrow$ C6 $\rightarrow$ plug). Cross-model answer anchoring and drop-one region tests assess answer reliability and evidence dependence, followed by multiple-choice conversion and difficulty filtering. During training, the policy receives the full image with region outlines, while privileged crops are used only for data construction.}
    \label{fig:method}
    \end{figure*}

As discussed in Section~\ref{sec:intro}, fine-grained perception, spatial perception, and multimodal long-context understanding share a core capability that we call connecting evidence.
This motivates training data whose questions structurally require this capability and end in concise answers suitable for verifiable rewards.
To construct this supervision, we propose \ours{}, a data synthesis framework centered on coherent groups of image regions. Within this framework, high-resolution privileged crops render local details visible for question synthesis, while cross-model answer anchoring and region ablation check label reliability and evidence dependence.
The overview of the data synthesis pipeline is shown in Figure~\ref{fig:method}.

\subsection{Multi-region Question Synthesis}
\label{sec:method_evidence}

\paragraph{Target question structure.}
\label{sec:method_objective}
We first define the target question, whose answer depends on connecting one coherent group of regions in an image.
Formally, from a source image \(I\), we obtain localized region proposals \(R_I\) and select one evidence group
\(G=\{r_{i_1},\ldots,r_{i_k}\}\subseteq R_I, 3\leq k\leq 7.
\)
Restricting the evidence to a single image keeps training inexpensive, while several related regions of one image already require a model to locate, bind, and compose dispersed evidence.
Importantly, the regions are selected to jointly support one question \(q\), which assigns each of them a role, i.e., the fact it contributes: a value to be combined, a member of a closed candidate set, or a clue on which a later step depends.

Corresponding to these three kinds of roles, \(q\) connects the regions through one of three forms.
\emph{Aggregation} combines counts or other numeric facts from the regions, possibly through a short calculation.
\emph{Comparison and matching} treats the regions as the complete candidate set, so that an answer such as an outlier, an extremum, or a pairing can be identified only by considering all of them.
\emph{Dependency chains and compositions} build the answer in several steps, e.g., by using a clue to decide what to read next, fixing a viewpoint for interpreting the rest, or combining predicates from different regions.
In all three forms, the answer is determined only when the assigned roles are taken together.
These forms also tie our questions to the target tasks: spatial relations, including those in an agent's or object's own frame, parallel spatial perception and embodied spatial reasoning, while binding facts to regions referenced by outline colors and integrating them parallels how long-context understanding and multimodal agents handle facts tied to image or turn references.
To synthesize such questions, we first locate candidate regions, then group those that support one of the three forms, and finally compose questions over each selected group.

\paragraph{Grounded evidence discovery.}
Since multi-region questions need detail-rich scenes, we start from high-resolution, visually complex images.
For each image, a strong VLM judges whether it can support such questions and, if so, is prompted to propose concrete, clearly localizable regions, each with a bounding box and a short description, prioritizing regions that can be related to one another over isolated salient ones.
We then discard proposals that are too small, too large, or heavily overlapping.
As these model-generated descriptions may be wrong, they only guide grouping and question writing, whereas reference answers are later derived from the image itself.

\paragraph{Task-conditioned evidence grouping.}
\label{sec:method_grouping}
An arbitrary collection of these proposals, however, does not necessarily support a valid compositional question.
We therefore prompt a strong model with the image caption and region descriptions to propose several groups per image, each labeled with a task family (a specific question type such as anomaly detection) and a necessity mode (how removing a region should affect the answer).
We then discard invalid or duplicated groups and closed sets of incomparable objects, and keep groups with distinct task families first.
The task family only guides question writing, whereas each region's concrete role is fixed when a question is written.

\paragraph{Privileged evidence layout.}
\label{sec:method_realization}
Since fine-grained details are difficult to inspect at full-image scale, we construct for each group a privileged evidence layout $V_G$, which is accessible only during data construction.
Following the layout-preserving design of \citet{liu2025hide}, $V_G$ arranges native-resolution crops of the selected regions, with small context margins, on a compact canvas that removes irrelevant background while preserving their relative horizontal and vertical order.
In addition, each region is outlined in a distinct color that also marks it in the full image, allowing questions to refer to regions by color unambiguously.
Finally, $V_G$ serves as the sole visual input to the question generator, which confines its questions to evidence within the group, and it is later reused for verification (Section~\ref{sec:method_verification}).

\paragraph{Question realization.}
The question generator, a strong VLM, then receives \(V_G\), the region descriptions, and a few reasoning blueprints, i.e., templates of reasoning programs, and composes several candidate questions, each with a concise answer: a single value (e.g., a number, a short phrase, or the outline color of a selected region), a set, or an ordered pair.
To make its use of evidence checkable, the generator also states each region's contribution and the answer expected after removing it, together with a reasoning certificate, i.e., a short sequence of steps that links the regions to the answer.
Deterministic rules then reject candidates whose certificate leaves a region unused or contains steps that do not lead to the answer, whose own predicted answer survives the removal of a region, or that use subjective wording, leak the answer through an outline color, or contain inconsistent arithmetic.
Implementation details and prompt templates appear in Appendices~\ref{app:impl} and~\ref{app:prompts}.

\subsection{Two-stage Verification}
\label{sec:method_verification}

The checks above concern only the form of a candidate.
A compositional question, however, is useful supervision only if its reference answer is reliable and the intended evidence matters.
We assess these properties separately, since a plausible reasoning structure establishes neither.

\paragraph{Cross-model answer anchoring.}
To obtain a more reliable reference answer, two VLMs from different model families answer \(q\) from \(V_G\) several times, without seeing the region descriptions or the proposed answer.
A candidate is kept only if a clear majority of each model's responses supports the same answer and this answer becomes the reference \(a^\star\).

\paragraph{Adaptive drop-one testing.}
Agreement on \(a^\star\), however, does not show that the answer actually depends on the selected evidence.
Since every question is constructed to require all of its regions (Section~\ref{sec:method_objective}), an answer that survives the removal of a region indicates a defect, such as a decorative region or a shortcut that bypasses the intended reasoning.
To detect such defects, we mask each \(r_i\in G\) in the privileged layout, \(V_G^{-i}=\mathcal{M}_i(V_G)\), where \(\mathcal{M}_i\) replaces the crop of \(r_i\) with a gray placeholder.
A verifier from one anchoring family then answers \(q\) from this view, reporting insufficient information if the masked evidence is needed, and either such a report or a changed answer confirms that the question depends on \(r_i\).
If instead \(a^\star\) is still returned, or the response is unusable, we query the verifier again and mark \(r_i\) as \emph{empirically redundant} only if most queries recover \(a^\star\).
Note that a single redundant region, however, may reflect noise rather than a defect, because part of a masked region can remain visible in an adjacent crop through overlapping context margins, and individual responses also vary.
We therefore reject a question only when more than one region is redundant, which also improves synthesis efficiency, since a stricter rule would discard many potentially valid questions merely because one region was judged redundant.
The implementation details are shown in Appendix~\ref{app:impl}.

\subsection{Data Finalization}
\label{sec:method_training}

\paragraph{From privileged crops to the policy view.}
Since the policy, i.e., the model under training, never sees \(V_G\), each verified candidate is paired with a policy view \(V_I\), by default the source image with each selected region outlined in its assigned color.
To avoid ambiguous references in the full image, these outlines mark which instances a question refers to, without revealing the answer-bearing content that the policy must still perceive and compose.
For aggregate questions, whose answers do not depend on the outlines, we also try to replace marker references with natural descriptions, so that the policy must locate the evidence itself.
This unmarked version replaces the marked one only if independent audits by both anchoring families, without access to \(a^\star\), confirm unique references, an unchanged multi-region task grounded in the intended regions, and the answer \(a^\star\).

\paragraph{Multiple-choice conversion and difficulty filtering.}
To support rule-based rewards in RLVR, we then turn each question into a four-option multiple-choice question, adding three distractors in the answer format of \(a^\star\) so that the format does not reveal the answer.
Each released example is thus \(x=(V_I,\,q_{\mathrm{rel}},\,O)\) with label \(y\in\{\mathrm{A},\mathrm{B},\mathrm{C},\mathrm{D}\}\), where \(q_{\mathrm{rel}}\) is the verified question stem, rewritten without marker references when the unmarked view is used, \(O\) the shuffled option set, and \(y\) the option of \(a^\star\).
Finally, since overly easy examples provide little learning signal, a lightweight VLM answers each example several times, and we discard examples that it always answers correctly.

\raggedbottom
\section{Experiments}
\label{sec:experiments}

This section evaluates whether learning to connect evidence within a single image yields improvements that generalize across multimodal perception tasks.

\begin{table}[t]
    \centering
    \caption{Main results on 25 benchmarks (accuracy, \%).
    Domain averages and Overall are unweighted means within each domain
    and over all 25 benchmarks, respectively.}
    \label{tab:main_results_large}
    \begingroup
    \fontsize{7.5}{8.4}\selectfont
    \setlength{\tabcolsep}{4.8pt}
    \renewcommand{\arraystretch}{1.0}
    \setlength{\aboverulesep}{1.4pt}
    \setlength{\belowrulesep}{1.4pt}
    \begin{tabular}{>{\columncolor{gray!18}}cl*{9}{c}}
        \toprule
        \multicolumn{1}{c}{\multirow{2}{*}{\textbf{Domains}}}
        & \multirow{2}{*}{\textbf{Benchmarks}}
        & \multicolumn{3}{c}{\cellcolor{gray!10}\textbf{Base models}}
        & \multicolumn{3}{c}{\cellcolor{cyan!18}\textbf{ZwZ2}}
        & \multicolumn{3}{c}{\textbf{Advanced models}} \\
        \cmidrule(lr){3-5}\cmidrule(lr){6-8}\cmidrule(l){9-11}
        \multicolumn{1}{c}{}
        & & \cellcolor{gray!10}4B & \cellcolor{gray!10}9B & \cellcolor{gray!10}27B
        & \cellcolor{cyan!18}4B
        & \cellcolor{cyan!18}9B
        & \cellcolor{cyan!18}27B
        & GPT & GLM & DeepSeek \\
        \midrule
        & VStar & 82.20 & 83.77 & 87.96 & 89.00 & 90.58 & 91.62 & 75.92 & 83.77 & 78.53 \\
        & HR-4K & 82.37 & 84.88 & 84.25 & 88.62 & 89.75 & 89.88 & 83.87 & 84.25 & 79.13 \\
        & HR-8K & 78.37 & 82.50 & 80.00 & 83.63 & 87.62 & 88.00 & 79.13 & 80.63 & 75.00 \\
        & MME-en & 70.60 & 71.00 & 72.62 & 75.48 & 75.33 & 77.41 & 58.50 & 67.88 & 50.75 \\
        & MME-cn & 69.33 & 69.53 & 68.68 & 73.89 & 73.45 & 74.97 & 56.90 & 68.31 & 50.04 \\
        & ZoomBench & 50.41 & 49.57 & 53.61 & 58.22 & 59.53 & 61.78 & 54.44 & 56.09 & 49.94 \\
        & Chartography & 5.00 & 7.00 & 8.00 & 14.00 & 13.00 & 12.00 & 17.00 & 9.00 & 10.00 \\
        & PerceptionBench & 35.53 & 38.70 & 41.57 & 37.80 & 38.89 & 44.93 & 48.77 & 44.78 & 44.00 \\
        \rowcolor{green!10}
        \cellcolor{gray!18}\multirow{-9}{*}{\shortstack{Fine-grained\\perception}}
        & \textbf{Avg.} & 59.23 & 60.87 & 62.09 & \textbf{65.08} & \textbf{66.02} & \textbf{67.57} & 59.32 & 61.84 & 54.67 \\
        \midrule
        & TreeBench & 50.37 & 46.67 & 49.63 & 51.36 & 51.60 & 56.05 & 52.35 & 57.53 & 61.98 \\
        & O3-Bench & 53.91 & 59.13 & 65.80 & 63.77 & 67.83 & 72.46 & 66.38 & 62.32 & 73.62 \\
        & CountQA & 36.91 & 37.63 & 46.73 & 47.19 & 48.43 & 54.32 & 51.83 & 49.35 & 64.07 \\
        & MIMIC & 55.01 & 59.39 & 55.95 & 66.10 & 64.98 & 65.16 & 57.43 & 73.46 & 68.55 \\
        & MMSI-Bench & 30.70 & 32.70 & 33.50 & 32.30 & 33.90 & 37.10 & 32.80 & 37.20 & 34.60 \\
        & SPAR-Bench & 41.47 & 43.53 & 46.68 & 45.69 & 47.23 & 50.78 & 43.92 & 52.13 & 40.66 \\
        \rowcolor{green!10}
        \cellcolor{gray!18}\multirow{-7}{*}{\shortstack{Spatial\\perception}}
        & \textbf{Avg.} & 44.73 & 46.51 & 49.72 & \textbf{51.07} & \textbf{52.33} & \textbf{55.98} & 50.79 & 55.33 & 57.25 \\
        \midrule
        & MMLongBench$^{*}$ & 63.50 & 69.89 & 73.41 & 68.34 & 73.45 & 76.73 & 67.79 & 83.15 & 82.20 \\
        & MileBench & 60.08 & 68.96 & 71.97 & 70.08 & 71.46 & 73.94 & 63.34 & 71.33 & 65.79 \\
        & MemLens & 40.62 & 45.82 & 52.22 & 42.59 & 47.34 & 55.64 & 37.20 & 58.40 & 66.57 \\
        \rowcolor{green!10}
        \cellcolor{gray!18}\multirow{-4}{*}{\shortstack{Long-context\\understanding}}
        & \textbf{Avg.} & 54.73 & 61.56 & 65.87 & \textbf{60.34} & \textbf{64.08} & \textbf{68.77} & 56.11 & 70.96 & 71.52 \\
        \midrule
        & ERQA & 45.75 & 45.75 & 51.25 & 52.00 & 55.50 & 61.75 & 54.00 & 54.50 & 59.50 \\
        & ViewSpatial-Bench & 37.43 & 42.65 & 48.49 & 45.26 & 45.27 & 50.63 & 40.18 & 46.57 & 56.29 \\
        & MindCube-Tiny & 47.62 & 40.76 & 50.76 & 53.05 & 52.57 & 54.76 & 57.62 & 72.00 & 42.00 \\
        \rowcolor{green!10}
        \cellcolor{gray!18}\multirow{-4}{*}{\shortstack{Embodied\\spatial\\reasoning}}
        & \textbf{Avg.} & 43.60 & 43.05 & 50.17 & \textbf{50.10} & \textbf{51.11} & \textbf{55.71} & 50.60 & 57.69 & 52.60 \\
        \midrule
        & AgentVista & 12.44 & 8.61 & 12.92 & 17.22 & 14.83 & 18.18 & 21.53 & 22.01 & 9.09 \\
        & ScreenSpot-Pro & 60.09 & 63.88 & 70.21 & 61.73 & 66.98 & 70.59 & 63.25 & 40.48 & 54.40 \\
        & UI-Vision & 35.83 & 42.48 & 66.05 & 37.67 & 45.48 & 66.66 & 59.43 & 38.48 & 43.34 \\
        \rowcolor{green!10}
        \cellcolor{gray!18}\multirow{-4}{*}{\shortstack{Multimodal\\agent}}
        & \textbf{Avg.} & 36.12 & 38.32 & 49.73 & \textbf{38.87} & \textbf{42.43} & \textbf{51.81} & 48.07 & 33.66 & 35.61 \\
        \midrule
        & MMStar & 74.13 & 77.27 & 75.93 & 76.27 & 78.27 & 79.07 & 74.73 & 80.20 & 80.73 \\
        & BabyVision & 15.46 & 18.56 & 19.85 & 18.56 & 21.39 & 28.35 & 32.22 & 37.11 & 45.88 \\
        \rowcolor{green!10}
        \cellcolor{gray!18}\multirow{-3}{*}{\shortstack{General visual\\reasoning}}
        & \textbf{Avg.} & 44.80 & 47.92 & 47.89 & \textbf{47.42} & \textbf{49.83} & \textbf{53.71} & 53.48 & 58.66 & 63.31 \\
        \midrule
        \rowcolor{violet!14}
        \multicolumn{2}{l}{\textbf{Overall}}
        & 49.41 & 51.63 & 55.52
        & \textbf{54.79} & \textbf{56.59} & \textbf{60.51}
        & 54.02 & 57.24 & 55.47 \\
        \bottomrule
    \end{tabular}
    \par\vspace{2pt}
    \begin{minipage}{\linewidth}
        \fontsize{7}{8}\selectfont
        \textit{Notes.} Base models are Qwen3.5-4B/9B and Qwen3.6-27B.
        GPT: GPT-5.6-Luna; GLM: GLM-5.3-Flash.
        DeepSeek: DeepSeek-V4.1-Flash.
        All models are evaluated in low-thinking or non-thinking reasoning modes. The input resolution of the three advanced models is constrained by platform limitations.
        $^{\ast}$: extracted MMLongBench visual-core subset, including V-RAG,
        Image MM-NIAH, and VH-Multi (25 non-Counting tasks in total).
    \end{minipage}
    \endgroup
\end{table}

\subsection{Experimental Settings}

\textbf{Benchmarks.}~~
We evaluate on 25 benchmarks in six domains (Table~\ref{tab:main_results_large}).
\emph{(i) Fine-grained perception:} VStar~\citep{wu2024vstar}, HR-Bench 4K/8K~\citep{wang2025hrbench}, MME-RealWorld (EN/CN)~\citep{zhang2025mmerealworld}, ZoomBench~\citep{wei2026zooming}, Chartography~\citep{garre2026chartography}, and PerceptionBench~\citep{lin2026perceptionbench}.
\emph{(ii) Spatial perception:} TreeBench~\citep{wang2025treebench}, O3-Bench~\citep{li2025insight}, CountQA~\citep{tamarapalli2025countqa}, MIMIC~\citep{das2026more}, MMSI-Bench~\citep{yang2026mmsi}, and SPAR-Bench~\citep{zhang2026flatland}.
\emph{(iii) Multimodal long-context understanding:} MMLongBench~\citep{wang2025mmlongbench}, MileBench~\citep{song2024milebench}, and MemLens~\citep{ren2026memlens}.
\emph{(iv) Embodied spatial reasoning:} ERQA~\citep{erqa2025}, ViewSpatial-Bench~\citep{li2025viewspatial}, and MindCube-Tiny~\citep{wang2506mindcube}.
\emph{(v) Multimodal agents:} AgentVista~\citep{su2026agentvista}, together with ScreenSpot-Pro~\citep{li2025screenspot} and UI-Vision~\citep{nayak2025uivisiondesktopcentricguibenchmark} for GUI grounding.
\emph{(vi) General visual reasoning:} MMStar~\citep{chen2024we} and BabyVision~\citep{chen2026babyvision}.
We report accuracy (\%) on all benchmarks.

\textbf{Training.}~~
We train Qwen3.5-4B/9B~\citep{qwen3.5} and Qwen3.6-27B~\citep{qwen3.6-27b} in non-thinking mode with GRPO~\citep{shao2024deepseekmath} on the 20.4K examples synthesized by \ours{} (Appendix~\ref{sec:data_preparation}), without supervised fine-tuning.
All three scales share the same data and training configuration (Appendix~\ref{app:training}), and we refer to the resulting models as ZwZ2-4B/9B/27B.

\textbf{Baselines.}~~
We compare with two groups of baselines.
As references for absolute performance, Table~\ref{tab:main_results_large} includes three advanced models: GPT-5.6-Luna, GLM-5.3-Flash, and DeepSeek-V4.1-Flash.
To compare \ours{} with existing synthetic training data, we further train Qwen3.5-4B with the same GRPO recipe on the datasets of ZwZ~\citep{wei2026zooming}, TVG~\citep{zhang2026tvg}, F$^{2}$RVLM~\citep{bi2026f2rvlm}, CRIT~\citep{sung2026crit}, and MMHops~\citep{zhang2026mmhops} (Table~\ref{tab:perception12_core_4b}).
For CRIT, whose 84,199 training samples contain 269,467 QA pairs from GQA, ActivityNet Captions, and SPIQA in multi-image, interleaved image--text, and multi-turn formats, we uniformly sample about 30K QA pairs across the three domains and train on them without CoT supervision or additional annotations.

\subsection{Main Results}

Table~\ref{tab:main_results_large} shows that ZwZ2 improves over its base model on all 25 benchmarks at every scale, raising the overall average by 5.39, 4.96, and 4.99 points at 4B, 9B, and 27B, respectively.
In particular, fine-grained and spatial perception improve by 5.2--6.3 points on average, with gains of up to 11.1 points on MIMIC and 10.8 points on CountQA, and embodied spatial reasoning improves by 5.5--8.1 points, including 10.5 points on ERQA at 27B.
Qualitative examples are provided in Appendix~\ref{app:case_study}.

Notably, although ZwZ2 is trained only on single images, its gains carry over to multi-image and interactive settings.
Long-context understanding improves at every scale, most markedly at 4B, where the domain average rises by 5.6 points, including a gain of 10.0 points on MileBench.
Likewise, the multimodal agent benchmark AgentVista improves by 4.8--6.2 points, and GUI grounding also improves consistently.
These results support our premise that connecting evidence within an image transfers to tasks whose evidence spans images, dialogue turns, or interaction steps.

ZwZ2 remains competitive with frontier models. ZwZ2-27B achieves the highest overall accuracy (60.51\%), compared with 54.02\%, 57.24\%, and 55.47\% for GPT-5.6-Luna, GLM-5.3-Flash, and DeepSeek-V4.1-Flash, respectively, and leads the fine-grained perception domain.

\begin{table}[t]
   \centering
   \caption{Training Qwen3.5-4B on different synthetic datasets (accuracy, \%). Parentheses give the number of training examples. HR denotes HR-Bench; Chart and O3 denote Chartography and O3-Bench. MMSI, MindCube, and ViewSpatial denote MMSI-Bench, MindCube-Tiny, and ViewSpatial-Bench, respectively.}
   \label{tab:perception12_core_4b}
   \begingroup
   \fontsize{7.5}{9.0}\selectfont
   \renewcommand{\arraystretch}{1.0}
   \setlength{\tabcolsep}{1.25pt}
   \setlength{\aboverulesep}{1.8pt}
   \setlength{\belowrulesep}{1.8pt}
   \resizebox{\linewidth}{!}{%
   \begin{tabular}{l*{11}{c}>{\columncolor{green!10}}c}
       \toprule
        \multicolumn{1}{l}{\multirow{2}{*}{\textbf{Models}}}
        & \multicolumn{3}{c}{\cellcolor{gray!10}\textbf{\shortstack{Fine-grained perception}}}
        & \multicolumn{3}{c}{\cellcolor{gray!10}\textbf{\shortstack{Spatial perception}}}
        & \multicolumn{3}{c}{\cellcolor{gray!10}\textbf{\shortstack{Embodied spatial reasoning}}}
        & \multicolumn{2}{c}{\cellcolor{gray!10}\textbf{\shortstack{Long-context understanding}}}
        & \multicolumn{1}{c}{\multirow{2}{*}{\textbf{Avg.}}} \\
        \cmidrule(lr){2-4}
        \cmidrule(lr){5-7}
        \cmidrule(lr){8-10}
        \cmidrule(lr){11-12}
        \multicolumn{1}{c}{}
        & \textbf{HR-4K}
        & \textbf{HR-8K}
        & \textbf{Chart}
        & \textbf{O3}
        & \textbf{CountQA}
        & \textbf{MMSI}
        & \textbf{MindCube}
        & \textbf{ERQA}
        & \textbf{ViewSpatial}
        & \textbf{MileBench}
        & \textbf{MemLens}
        & \multicolumn{1}{c}{} \\
        \midrule
       Base & 82.37 & 78.37 & 5.00 & 53.91 & 36.91 & 30.70 & 47.62 & 45.75 & 37.43 & 60.08 & 40.62 & \cellcolor{green!10}47.16 \\
       MMHops (31K) & 77.75 & 73.12 & 3.00 & 45.80 & 30.82 & 30.50 & 42.10 & 43.75 & 40.90 & 66.29 & 15.97 & \cellcolor{green!10}42.73 \\
       TVG (19.4K) & 78.25 & 73.62 & 13.00 & 53.04 & 33.97 & 29.50 & 28.86 & 46.00 & 42.16 & 64.56 & 37.29 & \cellcolor{green!10}45.48 \\
       F$^{2}$RVLM (37K) & 80.13 & 77.62 & 10.00 & 54.20 & 32.72 & 31.90 & 34.38 & 44.75 & 40.56 & 64.58 & 39.80 & \cellcolor{green!10}46.42 \\
       CRIT (30K) & 82.25 & 77.88 & 6.00 & 45.22 & 30.56 & 29.90 & 43.43 & 45.00 & 43.98 & 68.13 & 40.40 & \cellcolor{green!10}46.61 \\
       ZwZ (73K) & 87.00 & 82.75 & 7.00 & 60.87 & 43.32 & 31.33 & 33.81 & 46.00 & 43.75 & 70.21 & 42.46 & \cellcolor{green!10}49.86 \\
       \midrule
       \rowcolor{blue!5}
       \textbf{Ours (20.4K)} & 88.62 & 83.63 & 14.00 & 63.77 & 47.19 & 32.30 & 53.05 & 52.00 & 45.26 & 70.08 & 42.59 & \cellcolor{green!10}\textbf{53.86} \\
       \bottomrule
   \end{tabular}%
   }
   \endgroup
\end{table}

\subsection{Comparison with Existing Training Data}
\label{sec:exp_ablation}

Table~\ref{tab:perception12_core_4b} compares \ours{} with existing synthetic datasets on the same 4B base model.
These datasets target single-region fine-grained perception (ZwZ), grounded counting and spatial reasoning (TVG), dialogue-fragment retrieval (F$^{2}$RVLM), cross-modal multi-hop QA over interleaved contexts (CRIT), and knowledge-based multi-hop QA (MMHops).
Training on TVG, F$^{2}$RVLM, CRIT, or MMHops improves MileBench but degrades HR-Bench, CountQA, MindCube-Tiny, and MemLens, and all four datasets lower the average accuracy below that of the base model.
In contrast, \ours{} improves every benchmark over the base model and outperforms these four datasets on all 11 benchmarks, with a comparable amount of data (20.4K examples versus 19.4K--37K).

ZwZ-73K is the strongest baseline.
Since it also synthesizes questions from zoomed-in views but grounds each question in a single region, the comparison with ZwZ serves as an approximate ablation of connecting multiple regions.
Although ZwZ uses 3.6$\times$ as many examples (73K versus 20.4K), \ours{} raises the average accuracy from 49.86 to 53.86 and surpasses ZwZ on 10 of the 11 benchmarks, trailing it only on MileBench by 0.13 points.
The largest margin appears on MindCube-Tiny, where ZwZ reduces accuracy by 13.8 points whereas \ours{} improves it by 5.4 points.
This contrast is consistent with our premise that embodied spatial reasoning relies on connecting evidence across regions rather than on inspecting a single region.

\subsection{Ablation Studies}

\begin{table}[t]
   \centering
   \caption{Data-synthesis ablations on Qwen3.5-4B (accuracy, \%). Each trained variant uses 7K examples. The 7K reference model is shared by both comparisons and is reported only once; it uses privileged views and drop-one verification. Avg. is the unweighted mean over 11 benchmarks.}
   \label{tab:synthesis_ablations}
   \label{tab:sa1b_alignment}
   \label{tab:drop_one_ablation}
   \begingroup
   \fontsize{7.5}{9.0}\selectfont
   \renewcommand{\arraystretch}{1.0}
   \setlength{\tabcolsep}{1.25pt}
   \setlength{\aboverulesep}{1.8pt}
   \setlength{\belowrulesep}{1.8pt}
   \resizebox{\linewidth}{!}{%
   \begin{tabular}{l*{11}{c}>{\columncolor{green!10}}c}
       \toprule
        \multicolumn{1}{l}{\multirow{2}{*}{\textbf{Settings}}}
        & \multicolumn{3}{c}{\cellcolor{gray!10}\textbf{\shortstack{Fine-grained perception}}}
        & \multicolumn{3}{c}{\cellcolor{gray!10}\textbf{\shortstack{Spatial perception}}}
        & \multicolumn{3}{c}{\cellcolor{gray!10}\textbf{\shortstack{Embodied spatial reasoning}}}
        & \multicolumn{2}{c}{\cellcolor{gray!10}\textbf{\shortstack{Long-context understanding}}}
        & \multicolumn{1}{c}{\multirow{2}{*}{\textbf{Avg.}}} \\
        \cmidrule(lr){2-4}
        \cmidrule(lr){5-7}
        \cmidrule(lr){8-10}
        \cmidrule(lr){11-12}
        \multicolumn{1}{c}{}
        & \textbf{HR-4K}
        & \textbf{HR-8K}
        & \textbf{Chart}
        & \textbf{O3}
        & \textbf{CountQA}
        & \textbf{MMSI}
        & \textbf{MindCube}
        & \textbf{ERQA}
        & \textbf{ViewSpatial}
        & \textbf{MileBench}
        & \textbf{MemLens}
        & \multicolumn{1}{c}{} \\
        \midrule
       Base & 82.37 & 78.37 & 5.00 & 53.91 & 36.91 & 30.70 & 47.62 & 45.75 & 37.43 & 60.08 & 40.62 & \cellcolor{green!10}47.16 \\
       \midrule
       Without privileged views & 85.12 & 82.50 & 13.00 & 53.62 & 40.51 & 33.00 & 50.19 & 52.25 & 43.87 & 70.22 & 40.87 & \cellcolor{green!10}51.38 \\
       Without drop-one & 85.75 & 79.75 & 8.00 & 58.26 & 41.30 & 34.10 & 47.14 & 51.50 & 43.84 & 70.16 & 41.48 & \cellcolor{green!10}51.03 \\
       \textbf{Ours (7K)} & 86.62 & 82.87 & 9.00 & 62.90 & 45.09 & 34.60 & 52.29 & 47.50 & 45.68 & 70.47 & 42.14 & \cellcolor{green!10}\textbf{52.65} \\
       \bottomrule
   \end{tabular}%
    }
    \endgroup
\end{table}

\paragraph{Ablation study on drop-one verification.}
To evaluate adaptive drop-one verification during data synthesis, we compare Qwen3.5-4B trained on 7K examples constructed with or without this verification step (Table~\ref{tab:drop_one_ablation}).
As described in Section~\ref{sec:method_verification}, this step masks one selected region at a time in the privileged layout and tests whether the reference answer remains recoverable.
When the verifier returns the reference answer or an unusable response, additional queries are used; a region is marked as empirically redundant only if most queries recover the reference answer.
A question is rejected only when more than one region is empirically redundant, allowing for noise from overlapping crop margins and variable verifier responses.
With drop-one verification, average accuracy rises from 51.03 to 52.65 (+1.63 points), with improvements on 10 of the 11 benchmarks.
The largest gains are on MindCube-Tiny (+5.15), O3-Bench (+4.64), and CountQA (+3.79), while ERQA decreases by 4.00 points.
These results support the usefulness of this synthesis-time verification step for downstream performance, although its gains are not uniform across tasks.

\paragraph{Ablation study on privileged views.}
To examine the role of privileged views, we train Qwen3.5-4B on 7K examples synthesized from privileged views or original images, with all other settings fixed (Table~\ref{tab:sa1b_alignment}).
Synthesis from privileged views achieves a higher average accuracy than synthesis from original images (52.65 versus 51.38), with higher scores on 9 of the 11 benchmarks.
Original-image synthesis performs better on Chartography and ERQA, by 4.00 and 4.75 points, respectively.
Thus, privileged-view synthesis improves average accuracy by 1.27 points in this 7K comparison, although its gains are not uniform across benchmarks.

\subsection{Analysis}

\newsavebox{\AnalysisFigureBox}
\newcount\AnalysisWrapLines
\sbox{\AnalysisFigureBox}{%
    \begin{minipage}[b]{0.50\linewidth}
\begin{minipage}[t]{0.485\linewidth}
        \vspace{0pt}
        \centering
        \captionsetup{font=footnotesize,skip=3pt}
        \includegraphics[width=\linewidth]{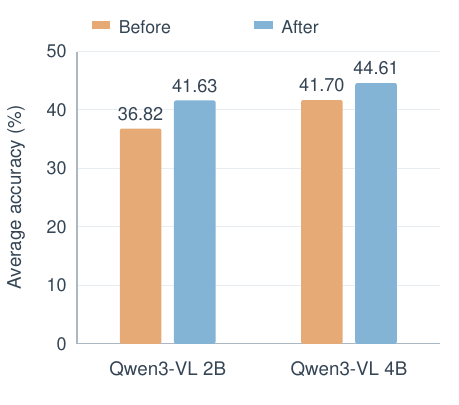}
        \captionof{figure}{Qwen3-VL generalization.}
        \label{fig:qwen3vl_generalization}
    \end{minipage}\hfill
    \begin{minipage}[t]{0.485\linewidth}
        \vspace{0pt}
        \centering
        \captionsetup{font=footnotesize,skip=3pt}
        \includegraphics[width=\linewidth]{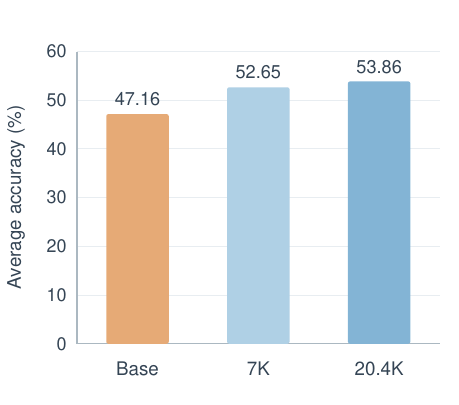}
        \captionof{figure}{Effect of training data scale.}
        \label{fig:data_scale}
    \end{minipage}
    \end{minipage}%
}
\begingroup
    \dimen0=\dimexpr\ht\AnalysisFigureBox+\dp\AnalysisFigureBox+6pt\relax
    \AnalysisWrapLines=\numexpr\dimen0/\dimexpr\baselineskip\relax+1\relax
    \ifdim\dimexpr\pagegoal-\pagetotal\relax<\dimen0
        \newpage
    \fi
    \hangindent=0.52\linewidth
    \hangafter=-\AnalysisWrapLines
    \noindent
    \makebox[0pt][l]{%
        \hspace*{-0.52\linewidth}%
        \raisebox{\dimexpr\ht\strutbox-\ht\AnalysisFigureBox\relax}[0pt][0pt]{\usebox{\AnalysisFigureBox}}%
    }%
\textbf{Generalization to Qwen3-VL.}
To examine whether the gains extend beyond the base models used in the experiments, we evaluate Qwen3-VL-2B and Qwen3-VL-4B before and after training on the same 11 benchmarks in Table~\ref{tab:perception12_core_4b}.
Figure~\ref{fig:qwen3vl_generalization} reports average scores of 36.82 and 41.63 for Qwen3-VL-2B, and 41.70 and 44.61 for Qwen3-VL-4B, corresponding to gains of 4.81 and 2.91 points, respectively.
These results suggest that the benefits of supervision extend to additional base models at both scales.
    \par
    \advance\AnalysisWrapLines by -\prevgraf
    \ifnum\AnalysisWrapLines>0
        \hangindent=0.52\linewidth
        \hangafter=-\AnalysisWrapLines
    \fi
    \vspace{3pt}
    \noindent
\textbf{Effect of training data scale.}
Figure~\ref{fig:data_scale} compares Qwen3.5-4B at different training data scales on the same 11 benchmarks.
Average accuracy over the 11 benchmarks in Table~\ref{tab:perception12_core_4b} increases from 47.16 for the base model to 52.65 with 7K training examples and 53.86 with the full 20.4K training set.
The full set yields a further improvement of 1.21 points over 7K examples, while the 7K setting already improves over the base model by 5.49 points.
    \par
    \advance\AnalysisWrapLines by -\prevgraf
    \ifnum\AnalysisWrapLines>0
        \vspace*{\dimexpr\AnalysisWrapLines\baselineskip\relax}
    \fi
\endgroup
\par\smallskip

\begin{figure*}[!htbp]
    \centering
    \captionsetup{font=small}
    \begin{minipage}[t]{0.655\textwidth}
        \vspace{0pt}
        \captionsetup{justification=raggedright,singlelinecheck=false}
        \includegraphics[width=0.485\linewidth]{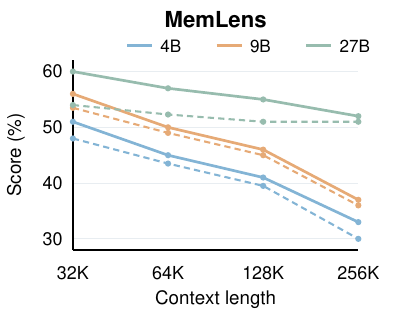}\hfill
        \includegraphics[width=0.485\linewidth]{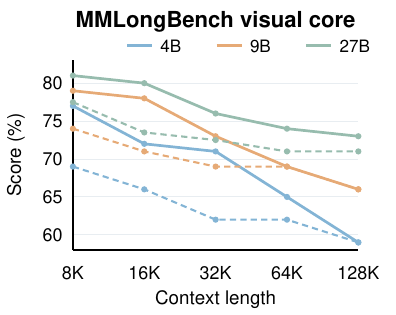}
        \captionof{figure}{Context-length performance on MemLens (left) and the MMLongBench visual core (right). Blue, orange, and green denote 4B, 9B, and 27B. Dashed and solid lines show the base models and ZwZ2, respectively.}
        \label{fig:context_length_combined}
    \end{minipage}\hfill
    \begin{minipage}[t]{0.325\textwidth}
        \vspace{0pt}
        \includegraphics[width=\linewidth]{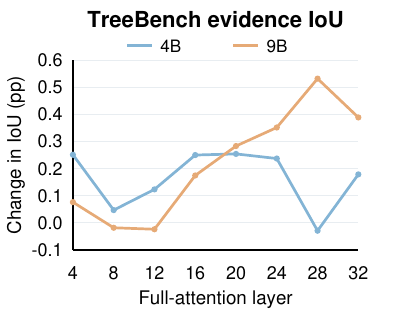}
        \captionof{figure}{\textbf{Attention alignment on TreeBench.} Layer-wise change in mean top-$k$ IoU (ZwZ2 minus Qwen3.5, percentage points).}
        \label{fig:treebench_attention_iou}
    \end{minipage}
\end{figure*}

\paragraph{Context length.}
Figure~\ref{fig:context_length_combined} breaks down MemLens and the MMLongBench visual core by context length.
ZwZ2 outperforms or matches its base model at every length and scale, and on MemLens its gains persist up to 256K tokens.
The gains, however, narrow at the longest contexts.
At 128K on the MMLongBench visual core, for example, ZwZ2-4B and ZwZ2-9B merely match their base models.
This trend is consistent with our training data, in which each example contains a single image and thus a short context.
Training therefore strengthens evidence connection but not retrieval from very long inputs, which likely becomes the main difficulty at the longest lengths.

\paragraph{Attention to evidence.}
\label{sec:attention_coverage}
To examine whether ZwZ2 focuses more on task-relevant visual evidence, we follow \citet{wei2026zooming} and evaluate ZwZ2-4B/9B against their base models on 405 TreeBench~\citep{wang2025treebench} examples equipped with annotated evidence bounding boxes.
For each full-attention layer, we compute the Intersection over Union (IoU) between the ground-truth evidence tokens $E$ and the $\vert{}E\vert{}$ visual tokens with the highest attention weights (Appendix~\ref{app:attention_coverage}).
As shown in Figure~\ref{fig:treebench_attention_iou}, ZwZ2 achieves a higher mean IoU across 7 out of 8 layers at the 4B scale and 6 out of 8 layers at the 9B scale.
This indicates that the high-attention tokens of ZwZ2 align more closely with the ground-truth evidence in most layers at both scales.

\section{Conclusion}
\label{sec:conclusion}

We presented \ours{}, a fully automated framework that synthesizes multi-region visual questions and answers, requiring a model to connect evidence within an image.
Trained with RLVR on these data alone, ZwZ2 improves over its base models on all 25 benchmarks at 4B, 9B, and 27B.
Notably, although training uses only single images, the gains extend beyond fine-grained and spatial perception to embodied spatial reasoning, multimodal long-context understanding, and multimodal agent tasks.
These results suggest that connecting evidence is a transferable training target, and thus tasks whose training data are costly to build can benefit from data that are far easier to construct and verify.

\section*{Acknowledgement}
Weiran Huang is supported by National Natural Science Foundation of China (No. 62406192), Shanghai Municipal Special Program for Basic Research on General AI Foundation Models (Grant No. 2025SHZDZX025G03), Tencent WeChat Rhino-Bird Focused Research Program, and Kuaishou Technology.
This work is supported by Ant Group Research Intern Program.
This work is also supported by the Zhongguancun Academy (Grant No.s C20250203).
We also sincerely thank Xiangyu Zou (Z.ai) for assistance with a portion of the model evaluations.

\bibliography{iclr2027_conference}
\bibliographystyle{iclr2027_conference}

\clearpage
\appendix
\onecolumn
\begin{center}
  \Large \textbf{Appendix}\\[0.5cm]
\end{center}

\section{Implementation Details}
\label{app:impl}

This section complements Section~\ref{sec:method} with the concrete settings used to construct our training data, following the order of that section.
The pipeline mainly uses Qwen3.5-397B-A17B~\citep{qwen3.5}, and Table~\ref{tab:impl_models} summarizes all model calls, each with the thinking mode disabled.

\begin{table}[h]
  \centering
  \caption{Model calls in the \ours{} pipeline. \(V_G\) denotes the privileged layout, \(V_G^{-i}\) the layout with region \(r_i\) masked, and \(V_I\) the policy view.}
  \label{tab:impl_models}
  \small
  \resizebox{\linewidth}{!}{%
  \begin{tabular}{lllll}
      \toprule
      \textbf{Stage} & \textbf{Model} & \textbf{Visual input} & \textbf{Calls} & \textbf{Temp.} \\
      \midrule
      Evidence discovery    & Qwen3.5-397B-A17B            & source image      & 1 per image          & 0.4 \\
      Evidence grouping     & Qwen3.5-397B-A17B            & none (text only)  & 1 per image          & 0.3 \\
      Question realization  & Qwen3.5-397B-A17B            & \(V_G\)           & 1 per group          & 0.5 \\
      Answer anchoring      & Qwen3.5-397B-A17B, Kimi-K2.6 & \(V_G\)           & up to 4 + 4 per candidate & 0.4 \\
      Equivalence judging   & GLM-5.2                      & none (text only)  & when needed          & 0.0 \\
      Drop-one testing      & Qwen3.5-397B-A17B            & \(V_G^{-i}\)      & 1--3 per region      & 0.4 \\
      Unmarked rewrite      & Qwen3.5-397B-A17B            & unmarked source image      & 1 per aggregate question & 0.2 \\
      Unmarked audit        & Qwen3.5-397B-A17B, Kimi-K2.6 & source image      & 2 + 2 per rewrite    & 0.0 \\
      Distractor generation & Qwen3.5-397B-A17B            & marked image      & 1 per marked candidate & 0.8 \\
      Difficulty filtering  & Qwen3.6-35B-A3B              & \(V_I\)           & 4 per example        & 0.7 \\
      \bottomrule
  \end{tabular}%
  }
\end{table}

\paragraph{Image source and pre-filtering.}
We use SA-1B~\citep{kirillov2023segment} as the sole image source.
Before any model call, we keep only images whose longer side has at least 1280 pixels, whose mean per-channel intensity entropy is at least 6.5 bits, whose Canny edge density is at least 0.04, and whose luminance standard deviation is at least 60.

\paragraph{Grounded evidence discovery.}
For each image, a single call with the image and its caption (Prompt~\ref{app:prompt-region-inventory}) judges whether the image can support multi-region questions and returns up to 18 region proposals, whose box coordinates are converted to pixels and clamped to the image.
We then discard boxes that cover less than 0.04\% or more than 18\% of the image, as well as boxes whose IoU with an earlier kept box is at least 0.65.
Images judged unsuitable, or with fewer than four remaining regions, are skipped.

\paragraph{Task-conditioned evidence grouping.}
A text-only call receives the caption together with the label and local description of each region, ranked by grounding confidence (Prompt~\ref{app:prompt-group-proposal}), and proposes up to six groups of 3--7 regions.
Each group is annotated with one of 22 task families, which map to four necessity modes that refine the three question forms of Section~\ref{sec:method_objective}.
Specifically, \texttt{answer\_sensitive} covers aggregation (counting, numeric aggregation, and arithmetic), \texttt{set\_complete} (filtering, anomaly, extremum, matching, and geometric, topological, or depth relations) and \texttt{pair\_complete} (cross-region pairing) cover comparison and matching, and \texttt{chain\_dependency} (OCR, spatial, viewpoint, conditional, Boolean, relational, and multi-hop chains) covers dependency chains and compositions.
We keep groups whose region identifiers are valid, whose size lies in the allowed range, and that do not repeat an earlier group.
In addition, the candidates of a filtering, anomaly, or extremum group must not mix coarse semantic classes, and groups with distinct task families are kept before groups that repeat a family.

\paragraph{Privileged evidence layout.}
Each region is cropped at native resolution with a margin of 8\% of its box size on every side.
The crops are placed on a white canvas through an order-preserving remapping of each image axis, which removes the rows and columns not covered by any crop, and the canvas is uniformly downscaled only when its longer side exceeds 4096 pixels.
Each region receives a distinct color from a seven-color palette, which bounds the group size at seven.
The same colors are used in the question text, in the privileged and ablated layouts, and for the outlines on the marked policy view.

\paragraph{Question realization.}
For each group, the generator receives the privileged layout, the group metadata (labels, local descriptions, natural referents, box coordinates, and outline colors), four blueprints sampled from a library of 104 reasoning templates, and up to four additional free-form slots, and it returns up to eight candidates (Prompt~\ref{app:prompt-question-realization}).
The first blueprint follows the proposed task family when possible, and the remaining ones diversify answer forms and question types.
Each candidate declares its question type, necessity mode, answer semantics (scalar, unordered set, or ordered pair), answer kind (outline color, number, Boolean pattern, semantic phrase, text, semantic set, or semantic pair), the contribution of every region, the predicted answer after removing each region, and a reasoning certificate.
Open-vocabulary answers must contain fewer than ten words.

\paragraph{Rule-based checks and candidate selection.}
A certificate is valid only if every step reads from earlier steps alone, every selected region lies on a path to the output step, and no step is disconnected from it.
Aggregation, arithmetic, and Boolean questions must further use an executable certificate with a restricted operator set.
For aggregation questions, every region must receive a positive value, and the values must sum to the answer without all being one, since the answer would otherwise equal the number of outlined regions.
Similarly, arithmetic expressions must reference every region and evaluate to the answer.
We further reject candidates that expose region identifiers, do not refer to the marked regions, use conditional wording without explicit branches or outside conditional types, count vague background targets, request more than one output for a scalar answer, omit an explicit coordinate frame for spatial types, leak the answer through an outline color, use subjective phrasing, or declare an answer form that the answer does not match.
Candidates whose own predicted answer after removing some region equals their proposed answer are rejected as well.
The remaining candidates are ranked by the generator's self-assessed naturalness, visual confidence, and multi-region scores, with bonuses for compositional types and non-numeric answers, and this ranking only determines which candidates proceed to verification.
Per group, we keep up to four top-ranked candidates, preferring distinct question types and allowing at most one numeric and two outline-color answers.
Per image, we keep up to eight candidates but always retain the best candidate of each question type.

\paragraph{Cross-model answer anchoring.}
Each candidate is answered on the privileged layout by both anchoring families, with up to four samples per family and without the region descriptions or the generator's answer (Prompt~\ref{app:prompt-privileged-answer}).
For closed-form questions, the prompt also lists the allowed answer labels, and we first draw two samples per family.
Such a candidate is rejected early unless at least three of these four responses agree and each family contributes at least one of them.
Responses are clustered by normalized string matching with numeric equivalence, and if the largest cluster has fewer than six members, the text-only judge re-clusters all responses by semantic equivalence (Prompt~\ref{app:prompt-equivalence}).
A candidate is accepted if the largest cluster is not an insufficiency response and contains at least three of the four responses from each family, and its answer then becomes \(a^\star\).
Finally, \(a^\star\) is re-checked against the answer-form constraints above.

\paragraph{Adaptive drop-one testing.}
For each selected region, its crop in the privileged layout is replaced by a gray block with a red cross, while all other crops stay in place.
The first anchoring family answers each ablated view once, with a mode-specific addendum to the answering prompt (Prompt~\ref{app:prompt-drop-one}).
If the response matches \(a^\star\) or cannot be parsed, the same view is queried twice more, and the region is redundant if at least two of its three responses match \(a^\star\).
A candidate is rejected if two or more of its regions are redundant.

\paragraph{Policy view.}
Unmarked rewriting is attempted only for aggregate question types (object counts, attribute counts, and sums of visible numbers), whose answers do not depend on the outlines, whereas the answers of selection and pairing questions are often outline colors themselves.
The first anchoring family rewrites the question for the unmarked image and proposes distractors (Prompt~\ref{app:prompt-unmarked-rewrite}), and rewrites that still mention boxes or outlines are discarded.
Two audits from each family (Prompt~\ref{app:prompt-unmarked-audit}) must then all report unique references, task preservation, identifiable targets, multi-region dependence, and necessity of every evidence area.
Each audit must also recover \(a^\star\) and ground exactly one box per intended region, and the grounded boxes must match the intended regions one-to-one, where a grounded box matches a region if its center lies inside the region's box enlarged by 15\%.
Otherwise, the example keeps the marked policy view, on which the box of each selected region is outlined in its assigned color.

\paragraph{Multiple-choice conversion.}
For marked examples, the first anchoring family proposes three distractors given the marked image, the question, and \(a^\star\) (Prompt~\ref{app:prompt-mcq-conversion}), whereas unmarked examples reuse the distractors from the rewrite step.
We remove empty distractors, distractors that match \(a^\star\) after normalization, duplicates, and, for closed-form answers, options outside the label vocabulary.
If fewer than three distractors remain, we fill the missing options deterministically, for example with alternative subsets of set-valued answers, unused vocabulary labels, or neighboring integers for count answers.
The four options are then shuffled, and the prompt asks for the option letter.

\paragraph{Difficulty filtering.}
The screening model answers each example four times on its policy view (Prompt~\ref{app:prompt-difficulty}).
Examples with four correct responses or any unparseable response are discarded, and the remaining ones are labeled easy, medium, hard, or very hard when three, two, one, or none of the responses are correct.
Finally, we keep at most four examples per source image, and each training example consists of the policy view and the multiple-choice question.

\section{Prompt Templates for \texorpdfstring{\ours{}}{EvidenceConnect}}
\label{app:prompts}

\newcounter{prompt}[section]
\renewcommand{\theprompt}{\thesection.\arabic{prompt}}
\newcommand{\promptheading}[1]{%
\refstepcounter{prompt}\paragraph{Prompt \theprompt: #1}}

This section lists the prompt templates used by \ours{}.
We replace instance-specific values with angle-bracketed placeholders and condense long templates for readability.
Specifically, we keep their task definitions, key constraints, and main output fields, and omit examples, formatting reminders, repeated instructions, and fields unused in our setting.

\subsection{Evidence Discovery and Question Construction}
\label{app:prompts-construction}

\promptheading{Grounded region inventory}
\label{app:prompt-region-inventory}

\begin{verbatim}
[IMAGE]
You are a precise visual grounding annotator and data-quality gate.

Task: Decide whether this image can support natural questions whose
answer genuinely depends on 3--7 distinct regions, then propose up to
<MAX_REGIONS> useful evidence regions.

Image caption, if available: <IMAGE_CAPTION>

Return regions that are:
- visually concrete and clearly localizable;
- useful for multi-region reasoning, including people, animals, objects,
signs, screens, text regions, fine details, and distinctive parts;
- preferably small or medium-sized, spatially diverse, and not overly
overlapping;
- suitable for coherent tasks such as positive-contribution counting,
closed-set filtering, anomaly/extremum detection, candidate matching,
pairing, OCR--visual verification, or a short dependent spatial chain.

Prioritize groups of related regions over isolated interesting objects.
The image is suitable only if at least one coherent all-region task is
possible. Avoid vague regions such as "background," "scene," "sky,"
"room," or the whole image.

For each region, return:
- label: short name;
- bbox: [x1,y1,x2,y2] in pixels, or normalized 0--1000 coordinates;
- caption: one concise visible description;
- natural_referent: an identifying phrase without coordinates or IDs;
- role: candidate | reference | comparator | text_region |
      count_target | fine_part | distractor;
- confidence: grounding confidence.

Return JSON only:
{
"suitable_for_multi_region_vqa": true,
"richness_reason": "...",
"coherent_sets": [
  {"labels": [...], "task_family": "..."}
],
"regions": [
  {
    "label": "...", "bbox": [...], "caption": "...",
    "natural_referent": "...", "role": "...", "confidence": 0.0
  }
]
}
\end{verbatim}

\promptheading{Task-conditioned region-group proposal}
\label{app:prompt-group-proposal}

\begin{verbatim}
You are an expert VQA data curator. Below are <N> candidate evidence
regions detected in one image.

Image caption: <IMAGE_CAPTION>
Regions:
- <REGION_ID>: label=<LABEL>; caption=<LOCAL_CAPTION>
- ...

Propose up to <MAX_GROUPS> combinations that are good for natural
questions requiring many regions. A good combination is a coherent
candidate/evidence set, not a bag of unrelated boxes. Prefer the
largest valid set, but every selected region must have a real role.

Allowed task families and necessity modes:
- answer_sensitive: aggregate_count, aggregate_attribute_count,
aggregate_numeric, arithmetic_composition;
- set_complete: set_filtering, anomaly_detection, extremum,
candidate_matching, geometric_line_relation, topological_relation,
depth_ordering;
- pair_complete: cross_region_pairing;
- chain_dependency: ocr_visual_verification, spatial_chain,
dependent_chain, viewpoint_transform, agent_relative_direction,
spatial_transitivity, conditional_composition, boolean_composition,
relational_composition, multi_hop_reasoning.

Selection rules:
1. Select 3--7 regions and prefer larger groups only when coherent.
2. Every region must change an aggregate, belong to the closed
 candidate/pair set, or supply an indispensable chain constraint.
3. Prefer objective text, numbers, countable objects, repeated classes,
 comparable parts, and unambiguous attributes.
4. Filtering, anomaly, and extremum candidates must be comparable.
5. Every aggregate-count region must contribute a positive count.
6. Spatial spread is preferred, but coherence has priority.
7. Diversify task families and do not default to counting.
8. Skip the image if no valid combination exists.

Return JSON only:
[
{
  "region_ids": ["r000", "r001", "r003"],
  "task_family": "anomaly_detection",
  "necessity_mode": "set_complete",
  "reason": "short objective rationale"
}
]
\end{verbatim}

\promptheading{Privileged question and certificate realization}
\label{app:prompt-question-realization}

\begin{verbatim}
[PRIVILEGED LAYOUT WITH K COLOR-OUTLINED CROPS]
Create <NUM_QUESTIONS> diverse multi-region VQA questions.

Scope contract:
- The selected set is exactly <REGION_IDS>.
- Every question must genuinely use all K regions.
- evidence_region_ids, region_contributions, answer_if_dropped, and
the reasoning certificate must cover exactly this set.
- Never add decorative observations for otherwise unused regions.
- Questions must be answerable from visible pixels in this layout,
not from region captions.

Visible crop anchors (the same outline colors appear on the marked image):
- <REGION_ID>: label=<LABEL>; outline_color=<COLOR>; bbox=<BBOX>;
  caption=<LOCAL_CAPTION>; natural_referent=<REFERENT>
- ...

Task guidance:
- Preferred task family: <TASK_FAMILY>.
- Preferred necessity mode: <NECESSITY_MODE>.
- Selected reasoning blueprints: <TEMPLATE_SPECS>.
- Instantiate a blueprint only when its facts are objectively visible.
- Include diverse closed-set, comparison/matching, OCR/spatial/
relational, and compositional structures.
- Prefer select-then-report questions in which several regions identify
a target and a second visible property supplies the final answer.
- Keep the answer short, objective, unique, and absent from the stem.

Necessity contract:
- answer_sensitive: every region contributes a positive value, so
masking it changes the result or makes the aggregate unknown.
- set_complete / pair_complete: all regions form the explicitly closed
candidate or pairing set; masking one prevents proving completeness.
- chain_dependency: every region supplies an indispensable clue or hop.
- For every region, answer_if_dropped must differ from the reference
answer or be "insufficient information."

Certificate contract:
- First design a natural question from pixels; then provide its
certificate. Do not verbalize the certificate mechanically.
- Use mode="executable" for deterministic reading, counting, filtering,
matching, comparison, arithmetic, Boolean, and lookup programs.
- Use mode="evidence_graph" for open-vocabulary semantic or unusual
multi-hop reasoning.
- Every step has id, op, regions, inputs, and query.
- Inputs may reference only earlier steps.
- Every selected region must lie on a path to output_step.
- No disconnected or decorative step is allowed.

Wording contract:
- Ask for one final answer.
- Use one explicit coordinate frame for spatial questions.
- Use outline colors only as reference anchors, never as leaked object
attributes.
- Open answers must be objective and fewer than 10 words.
- Do not use subjective comparisons, vague counts, or invisible facts.

Return JSON only, one object per candidate:
[
{
  "template_id": "...",
  "question": "...",
  "question_type": "...",
  "necessity_mode": "answer_sensitive | set_complete |
                     pair_complete | chain_dependency",
  "answer_semantics": "scalar | unordered_set | ordered_pair",
  "final_answer_kind": "anchor | numeric | boolean | semantic |
                        semantic_set | semantic_pair | text",
  "evidence_region_ids": [...],
  "region_contributions": {"r000": "...", "...": "..."},
  "region_values": {"r000": 0},
  "aggregation": "...",
  "operation_expression": "...",
  "candidate_region_ids": [...],
  "closed_candidate_set": true,
  "pair_region_ids": [[...], ...],
  "chain_region_ids": [...],
  "coordinate_frame": "image_plane | camera | agent | object |
                       scene_3d | not_applicable",
  "answer_vocabulary": [...],
  "answer_if_dropped": {"r000": "...", "...": "..."},
  "hypothetical_answer": "...",
  "reasoning_certificate": {
    "mode": "executable | evidence_graph",
    "steps": [
      {"id": "s1", "op": "...", "regions": ["r000"],
       "inputs": [], "query": "..."}
    ],
    "output_step": "..."
  },
  "naturalness_score": 1,
  "visual_confidence": 1,
  "multi_region_score": 1
}
]
\end{verbatim}

\subsection{Answer Anchoring and Dependency Checks}
\label{app:prompts-verification}

\promptheading{Privileged-view answer anchoring}
\label{app:prompt-privileged-answer}

\begin{verbatim}
[PRIVILEGED LAYOUT]
Answer the question using only the provided image.

Question: <QUESTION>
Allowed canonical answer labels: <OPTIONAL_NEUTRAL_VOCABULARY>

You may reason briefly before answering. Put only the final short answer
inside <answer>...</answer>.

Rules:
- Use an objectively supported expression of fewer than 10 words.
- Use Arabic digits for numbers, with no units or separators.
- Do not emit coordinates, grounding tokens, or box markers.
- Use the same language as the question.
- If evidence is insufficient, output
<answer>insufficient information</answer>.
- The answer tag is required; responses without it are discarded.
\end{verbatim}

\promptheading{Semantic-equivalence clustering}
\label{app:prompt-equivalence}

\begin{verbatim}
You are an answer-equivalence judge for a VQA task.

Question: <QUESTION>
Answers:
A1: <ANSWER_1>
...
AN: <ANSWER_N>

Group answers by semantic equivalence. Equivalent paraphrases belong
together; answers with different values, attributes, or truth values
must remain separate.

Return strict JSON only:
{
"groups": [["<CANONICAL_FORM>", [<ANSWER_INDICES>]], ...],
"largest_group_size": <INTEGER>,
"canonical": "<CANONICAL_FORM_OF_LARGEST_GROUP>"
}

Every answer must occur in exactly one group. Break size ties using the
lexicographically smallest canonical form.
\end{verbatim}

\promptheading{Mode-specific drop-one verification}
\label{app:prompt-drop-one}

The drop-one verifier inserts one of the following addenda into
Prompt~\ref{app:prompt-privileged-answer}, depending on the necessity mode.

\begin{verbatim}
[answer_sensitive]
ABLATION VIEW: one required aggregate region has been replaced by a
gray masked box. If the full aggregate cannot be computed, answer
"insufficient information"; do not count only visible boxes.

[set_complete or pair_complete]
ABLATION VIEW: one colored candidate box has been replaced by a gray
masked box. The masked box remains a member of the explicitly closed
candidate/pairing set. If its hidden content prevents proving the
complete subset, unique anomaly, global extremum, unique match, or
complete pairing, answer "insufficient information."

[chain_dependency]
ABLATION VIEW: one required clue region has been replaced by a gray
masked box. If the missing clue breaks the reasoning chain, answer
"insufficient information"; do not guess.
\end{verbatim}

\subsection{Policy-View and Multiple-Choice Construction}
\label{app:prompts-release}

\promptheading{Unmarked-question rewrite}
\label{app:prompt-unmarked-rewrite}

\begin{verbatim}
[UNMARKED ORIGINAL IMAGE]
You are given an anchored visual question and its verified answer.

Anchored question: <QUESTION>
Verified answer: <ANSWER>

Create a minimally changed question that works on the unmarked image.
Replace box/outline references with visually unique natural descriptions
or a clearly exhaustive image-level scope. Do not add facts, change the
operation, or change the answer. If unique reference cannot be
guaranteed, set safe_without_marks=false. Also generate three plausible
incorrect distractors in the same answer format.

Return JSON only:
{
"safe_without_marks": true,
"natural_question": "...",
"distractors": ["...", "...", "..."]
}
\end{verbatim}

\promptheading{Independent audit of the unmarked rewrite}
\label{app:prompt-unmarked-audit}

\begin{verbatim}
[UNMARKED ORIGINAL IMAGE]
Audit this visual question using only the unmarked original image.

Anchored source question: <ANCHORED_QUESTION>
Unmarked natural question: <NATURAL_QUESTION>

Independently answer the question and judge whether every reference or
exhaustive target set is uniquely determined. The question must require
multiple spatially separated regions. Ground each evidence area with a
box in normalized 0--1000 coordinates, locating it from pixels rather
than inferring it from wording. Also judge whether the rewrite preserves
the source task and whether every grounded area is required.

Return JSON only:
{
"answer": "...",
"reference_status": "unique | ambiguous",
"same_task_as_source": true,
"all_targets_identifiable": true,
"requires_multiple_regions": true,
"every_evidence_area_required": true,
"evidence_boxes": [
  {"description": "...", "bbox_2d": [x1,y1,x2,y2]}
],
"reason": "..."
}
\end{verbatim}

\promptheading{Multiple-choice distractor generation}
\label{app:prompt-mcq-conversion}

\begin{verbatim}
[MARKED ORIGINAL IMAGE]
You are given a visual question and its correct answer.

Original question: <QUESTION>
Correct answer: <ANSWER>

Generate three plausible but incorrect multiple-choice distractors.
Do not rewrite, paraphrase, simplify, or neutralize the question.
Keep the stem exactly unchanged. Distractors must be visually relevant,
distinct from the answer, and matched in answer format and approximate
length. For open-vocabulary answers, use concrete alternatives with the
same grammatical form and specificity. Do not use "none of the above,"
"all of the above," or "cannot be determined."

Return JSON only:
[{"distractors": ["d1", "d2", "d3"]}]
\end{verbatim}

\promptheading{Difficulty sampling on the policy view}
\label{app:prompt-difficulty}

\begin{verbatim}
[POLICY VIEW: MARKED OR UNMARKED IMAGE]
Answer this multiple-choice visual question using only the image.

<QUESTION_AND_OPTIONS>

You may reason briefly before answering. The final answer must be:
<answer>X</answer>

Rules:
- X must contain only A, B, C, or D.
- Do not output option text, coordinates, grounding tokens, or markers.
- Keep any reasoning to a few short sentences.
- The answer tag is required; otherwise the response is discarded.
\end{verbatim}

\section{Training Data Statistics}
\label{sec:data_preparation}

\begin{figure}[t]
  \centering
  \includegraphics[width=\linewidth]{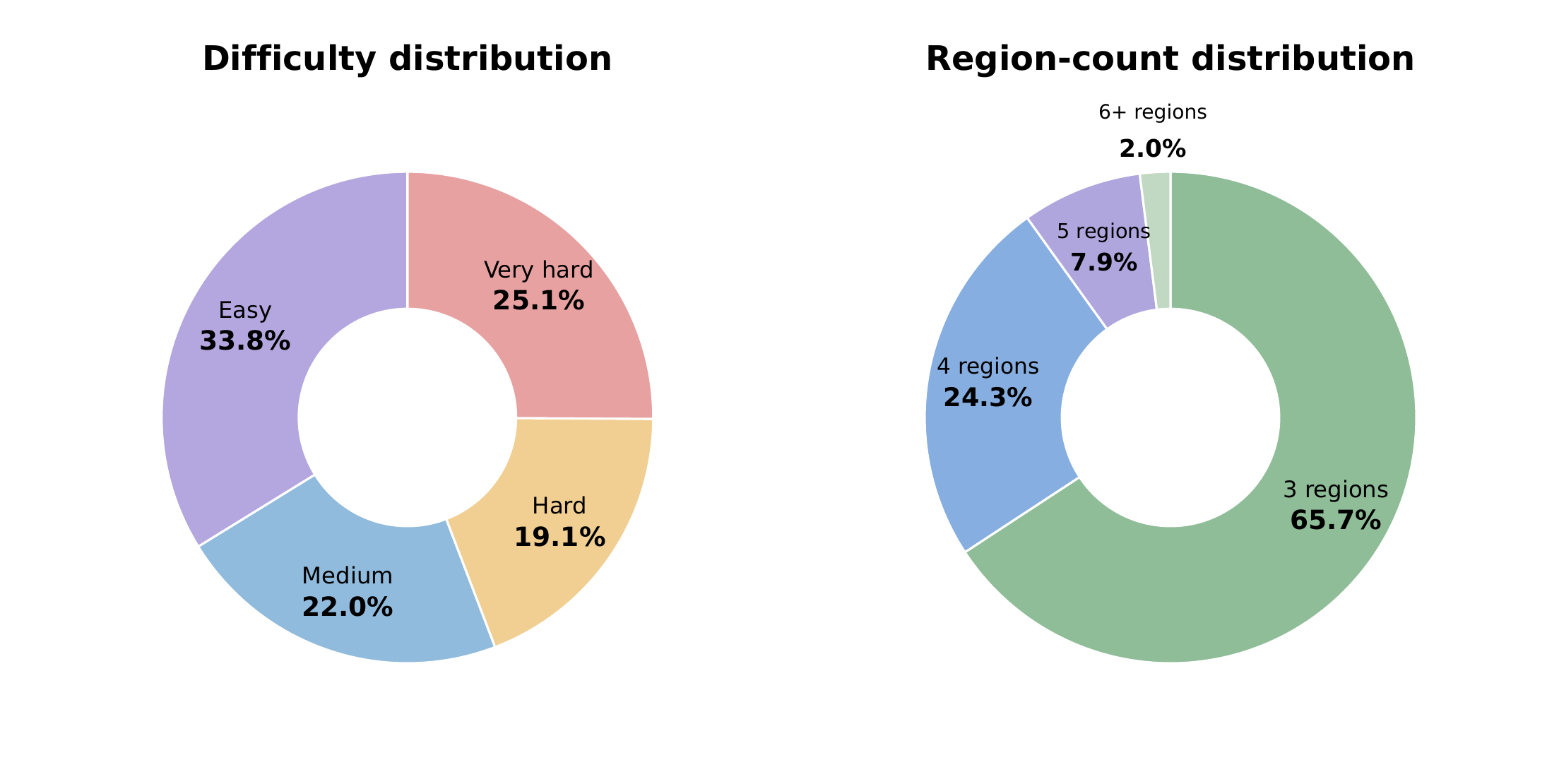}
  \caption{Composition of the 20.4K training examples by the number of selected regions per example and by difficulty level.}
  \label{fig:data_distribution}
\end{figure}

Our training set contains 20.4K examples synthesized from SA-1B images by the pipeline in Section~\ref{sec:method} and Appendix~\ref{app:impl}, and Figure~\ref{fig:data_distribution} summarizes its composition.
Each example involves three to seven regions, most often three (65.74\%) or four (24.34\%), followed by five (7.93\%) and six or seven (1.99\%).
The difficulty levels, defined by how many of the four screening responses are correct, are broadly distributed.
Specifically, easy, medium, hard, and very hard examples account for 33.75\%, 22.04\%, 19.07\%, and 25.14\% of the data, so more than 44\% of the examples are hard or very hard.
Very hard examples, which the screening model never answers correctly, are kept, since their reference answers have already been anchored by two stronger VLMs.

\section{Training Details}
\label{app:training}
All models are trained with the same GRPO~\citep{shao2024deepseekmath} configuration for two epochs, with a batch size of 128 and a learning rate of \(10^{-6}\).
For each prompt, we sample \(G{=}8\) rollouts at temperature 0.7 with a maximum response length of 32K tokens, and the thinking mode is disabled during rollouts to keep responses concise.
We use decoupled clipping ratios \(\epsilon_{\mathrm{low}}{=}0.2\) and \(\epsilon_{\mathrm{high}}{=}0.3\), as proposed in DAPO~\citep{yu2025dapo}, and remove the KL penalty (\(\beta_{\mathrm{KL}}{=}0\)), so that the policy is not anchored to the base model.
Finally, the reward checks whether the predicted option letter matches the ground-truth option.

\section{Attention Map Coverage Analysis}
\label{app:attention_coverage}

\paragraph{Evaluation setting.}
We compare ZwZ2-4B/9B with their Qwen3.5 base models on the same 405 TreeBench examples.
Within each scale, both checkpoints use the base-model processor and identical image--question pairs, visual-token grids, and evidence annotations.
Since Qwen3.5 interleaves linear-attention and full-attention layers, we evaluate all eight full-attention layers,
$\mathcal{L}=\{4,8,12,16,20,24,28,32\}$ (1-based).

\paragraph{Attention extraction.}
For each example, we apply the chat template with the generation prompt and perform a single forward pass in evaluation mode using eager attention, without generating an answer.
We extract attention from the final token of this complete input sequence to the image tokens.
This position predicts the first answer token, so its attention indicates where the model looks as it begins to answer.

Let $V_i$ denote the visual-token indices for example $i$, and let $q_i$ be its final input-token position.
For model $m$ at layer $\ell$, let
$T_{i,\ell,h}^{m}(q_i,v)$ denote the attention weight returned by head $h$ from $q_i$ to visual token $v$.
We convert the extracted weights to float32, average equally over all $n_{\mathrm{heads}}^{m,\ell}$ heads, and then renormalize within the image tokens:
\begin{equation}
\begin{aligned}
  u_{i,\ell}^{m}(v)
  &=
  \frac{1}{n_{\mathrm{heads}}^{m,\ell}}
  \sum_{h=1}^{n_{\mathrm{heads}}^{m,\ell}}
  T_{i,\ell,h}^{m}(q_i,v), \\
  a_{i,\ell}^{m}(v)
  &=
  \frac{u_{i,\ell}^{m}(v)}
  {\sum_{v'\in V_i}u_{i,\ell}^{m}(v')}.
\end{aligned}
\label{eq:attention_extraction}
\end{equation}
This renormalization over image tokens is applied once after head averaging, without renormalizing each head separately or applying an additional softmax.

\paragraph{Evidence mask.}
The visual tokens form an $H_i\times W_i$ grid for an image of size
$H_i^{\mathrm{img}}\times W_i^{\mathrm{img}}$.
An annotated box $(x_0,y_0,x_1,y_1)$ covers columns
$[\lfloor x_0W_i/W_i^{\mathrm{img}}\rfloor,
\lceil x_1W_i/W_i^{\mathrm{img}}\rceil)$
and the analogous row range, clipped to the grid boundaries.
This range includes every visual-token cell that overlaps the box.
The evidence set $E_i\subseteq V_i$ is the union of the tokens covered by all annotated boxes, counting overlapping tokens once.
All evaluated examples have $|E_i|>0$.

\paragraph{Top-$k$ evidence IoU.}
For each example, we set $k_i=|E_i|$ and select the $k_i$ visual tokens with the largest head-averaged weights $u_{i,\ell}^{m}(v)$ to form $A_{i,\ell}^{m}$.
Ranking by the normalized weights $a_{i,\ell}^{m}(v)$ gives the same ordering because the normalization divides all weights by a common positive value.
Ties are broken by visual-token index in row-major order, so exactly $k_i$ tokens are selected even when ties occur at the boundary.
Because $k_i$ is fixed by the annotation, the base and trained models receive the same token budget.

With $h_{i,\ell}^{m}=|A_{i,\ell}^{m}\cap E_i|$, we compute
\begin{equation}
  \operatorname{IoU}_{i,\ell}^{m}
  =\frac{|A_{i,\ell}^{m}\cap E_i|}
  {|A_{i,\ell}^{m}\cup E_i|}
  =\frac{h_{i,\ell}^{m}}{2k_i-h_{i,\ell}^{m}}.
  \label{eq:topk_evidence_iou}
\end{equation}
The score lies in $[0,1]$, with larger values indicating better spatial alignment with the annotated evidence.
Since $|A_{i,\ell}^{m}|=|E_i|$, precision equals recall,
$p=h_{i,\ell}^{m}/k_i$, and the IoU equals $p/(2-p)$.

\paragraph{Aggregation and layer-wise results.}
We take the unweighted mean of the per-example IoU over the $N=405$ paired examples, rather than pooling intersections and unions, and report differences in percentage points:
\begin{equation}
\begin{aligned}
  \overline{\operatorname{IoU}}_{\ell}^{m}
  &=
  \frac{1}{N}\sum_{i=1}^{N}\operatorname{IoU}_{i,\ell}^{m}, \\
  \Delta_{\ell}
  &=
  100\left(
  \overline{\operatorname{IoU}}_{\ell}^{\mathrm{ZwZ2}}
  -
  \overline{\operatorname{IoU}}_{\ell}^{\mathrm{Qwen3.5}}
  \right).
\end{aligned}
\end{equation}
Table~\ref{tab:treebench_attention_iou} shows gains in 7 of 8 layers at 4B and 6 of 8 layers at 9B.
The exceptions, layer 28 at 4B and layers 8 and 12 at 9B, each decrease by less than 0.03 percentage points.
These comparisons describe changes in sample means.
The metric measures spatial alignment at the final prompt position, rather than total attention mass or answer correctness.

\begin{table}[t]
  \centering
  \small
  \setlength{\tabcolsep}{7pt}
  \caption{\textbf{Top-$k$ evidence IoU on TreeBench.}
  Mean IoU is reported as a percentage; $\Delta$ is ZwZ2 minus Qwen3.5 in percentage points.
  Each entry averages the same 405 paired examples.
  Differences are computed before rounding.}
  \label{tab:treebench_attention_iou}
  \begin{tabular}{c rrr rrr}
      \toprule
      & \multicolumn{3}{c}{4B} & \multicolumn{3}{c}{9B} \\
      \cmidrule(lr){2-4}\cmidrule(lr){5-7}
      Layer & Qwen3.5 & ZwZ2 & $\Delta$
            & Qwen3.5 & ZwZ2 & $\Delta$ \\
      \midrule
      4  & 11.625 & 11.877 & +0.252 & 10.622 & 10.699 & +0.077 \\
      8  & 10.355 & 10.402 & +0.047 & 10.069 & 10.051 & -0.018 \\
      12 & 11.568 & 11.692 & +0.124 & 11.073 & 11.050 & -0.023 \\
      16 & 10.400 & 10.650 & +0.251 & 11.080 & 11.256 & +0.175 \\
      20 & 11.132 & 11.386 & +0.255 & 11.493 & 11.777 & +0.284 \\
      24 & 10.924 & 11.162 & +0.238 & 10.890 & 11.242 & +0.352 \\
      28 & 11.181 & 11.152 & -0.029 & 11.147 & 11.680 & +0.533 \\
      32 & 11.244 & 11.424 & +0.179 & 10.471 & 10.860 & +0.390 \\
      \bottomrule
  \end{tabular}
\end{table}

\section{Case Study}
\label{app:case_study}

\begin{figure}[t]
  \centering
  \includegraphics[width=\linewidth]{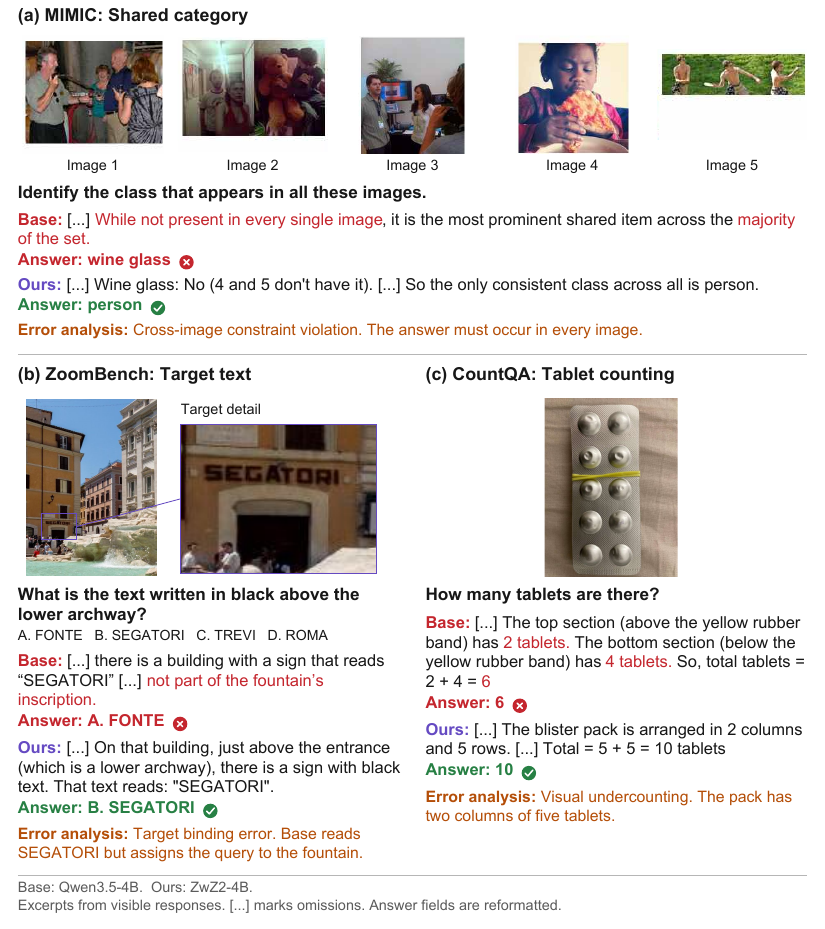}
  \caption{
      \textbf{Qualitative comparisons between Qwen3.5-4B and ZwZ2-4B.}
      Three selected examples show correct predictions by ZwZ2-4B where the base model fails:
      (a) identifying a category present in every image;
      (b) associating recognized text with the queried location;
      and (c) counting all instances in a structured arrangement.
      Responses are excerpted from the original outputs, with omissions marked by \texttt{[...]} and answer fields reformatted.
      The enlarged image region in (b) is provided for visualization only.
  }
  \label{fig:case_study}
\end{figure}

To complement the quantitative evaluation, we examine three selected examples in which ZwZ2-4B answers correctly while its base model, Qwen3.5-4B, fails.
Figure~\ref{fig:case_study} presents examples from MIMIC, ZoomBench, and CountQA, covering cross-image category identification, spatially specified text recognition, and object counting.
We analyze the models' visible responses to identify how their answers differ in the use of visual evidence and the satisfaction of question constraints.

\paragraph{Cross-image consistency.}
In Figure~\ref{fig:case_study}(a), the MIMIC question asks for a category that appears in all five images.
The base model selects \emph{wine glass}, even though its response explicitly acknowledges that this object is absent from some images.
It justifies the answer by referring to the most prominent shared item across the majority of the set, effectively replacing the requirement of universal presence with majority overlap.
ZwZ2 checks candidate categories against the full set of images.
Its response rules out \emph{wine glass} because it is absent from Images~4 and~5, and identifies \emph{person} as the category shared by all five images.
This example illustrates the importance of preserving the question's scope when integrating evidence across images: an object shared by several images does not necessarily satisfy a constraint over the entire set.

\paragraph{Question-conditioned text localization.}
Figure~\ref{fig:case_study}(b) asks for the black text above the lower archway.
The base model recognizes the word \emph{SEGATORI} on the adjacent building, but associates the question with inscriptions on the fountain and ultimately selects \emph{FONTE}.
Its response therefore reveals a mismatch between the recognized text and the region specified by the question.
ZwZ2 identifies the entrance of the yellow building as the relevant lower archway and reads the sign immediately above it, yielding the correct answer, \emph{SEGATORI}.
The case highlights the role of spatial reference resolution in visual text understanding: recognizing a visible word must be accompanied by establishing whether it belongs to the queried location.

\paragraph{Structured visual counting.}
In Figure~\ref{fig:case_study}(c), the CountQA question asks for the number of tablets in a blister pack.
The base model divides the image into regions above and below the yellow rubber band, reports two and four tablets, respectively, and returns $2+4=6$.
Its arithmetic is consistent with these counts, so the error lies in enumerating the tablets.
ZwZ2 describes the pack as two columns and five rows, counts five tablets in each column, and obtains the correct total of $5+5=10$.
Here, the response uses the regular spatial arrangement to account for the complete set of tablets.

Together, these cases reflect the steps of connecting evidence: integrating facts over the full set of images in (a), binding recognized text to the queried location in (b), and accounting for every instance before aggregating in (c).

\section{Limitations and Future Work}
\label{sec:limitations}
Our work has two main limitations.
First, although our single-image supervision transfers to multi-image and agent tasks, it does not directly model temporal or cross-frame dependencies, so extending the synthesis to evidence spread across video frames is a natural next step.
Second, we apply RLVR to models of up to 27B parameters, and whether the same supervision benefits larger models or earlier training stages, such as pre-training, remains to be studied.

\end{document}